\documentclass[journal,draftclsnofoot,onecolumn]{IEEEtran}

\usepackage{lineno,hyperref}
\modulolinenumbers[5]

\usepackage{url}
\usepackage{amsfonts}
\usepackage{amsmath}
\usepackage{amssymb}
\usepackage{graphicx} 
\usepackage{subfigure}
\usepackage{amsopn}
\usepackage{amsmath, bbold, esvect}
\usepackage[makeroom]{cancel}
\usepackage{amssymb,amsthm,pifont}
\usepackage{slashbox}

\newtheorem{lemma}{Lemma}

\newcommand\mum{\ensuremath{\mu\mathrm{m}}}

\begin{document}
%
\title{Wavelet-Based Semantic Features\\ for Hyperspectral Signature Discrimination}
%
%
%

\author{Siwei~Feng,~\IEEEmembership{Student~Member,~IEEE,}
Yuki~Itoh,~\IEEEmembership{Student~Member,~IEEE,}\\
Mario~Parente,~\IEEEmembership{Senior~Member,~IEEE,}
and~Marco~F.~Duarte,~\IEEEmembership{Senior~Member,~IEEE}
\thanks{The authors are with the Department
of Electrical and Computer Engineering, University of Massachusetts, Amherst,
MA, 01003, USA. E-mail: \{siwei, yitoh\}@engin.umass.edu and \{mparente, mduarte\}@ecs.umass.edu. Portions of this work have been presented at the IEEE Workshop on Hyperspectral Image and Signal Proc.: Evolution in Remote Sensing (WHISPERS)~\cite{MParente:a}, the Allerton Conference on Communication, Control, and Computing~\cite{MFDuarte:non-homogeneous} the IEEE Int. Conf. Image Processing (ICIP)~\cite{SFeng:tailoring}, and the IEEE Conference on Computer Vision and Pattern Recognition Workshops (CVPRW)~\cite{feng2015universality}.}
\thanks{This work was supported by the National Science Foundation under grant number IIS-1319585.}}

\maketitle

\begin{abstract}
Hyperspectral signature classification is a quantitative analysis approach for hyperspectral imagery which performs detection and classification of the constituent materials at the pixel level in the scene. The classification procedure can be operated directly on hyperspectral data or performed by using some features extracted from the corresponding hyperspectral signatures containing information like the signature's energy or shape. In this paper, we describe a technique that applies non-homogeneous hidden Markov chain (NHMC) models to hyperspectral signature classification. The basic idea is to use statistical models (such as NHMC) to characterize wavelet coefficients which capture the spectrum semantics (i.e., structural information) at multiple levels. Experimental results show that the approach based on NHMC models can outperform existing approaches relevant in classification tasks. 
\end{abstract}


\IEEEpeerreviewmaketitle

\section{Introduction}

Hyperspectral remote sensors collect reflected image data simultaneously in hundreds of narrow, adjacent spectral bands that make it possible to derive a continuous spectrum curve for each image cell. Such hyperspectral reflectance curves provide insight into the on-ground (or near ground) constituent materials in a single remotely sensed pixel. \par
The identification of ground materials from hyperspectral images often requires comparing the reflectance spectra of the image pixels, extracted endmembers, or ground cover exemplars to a training library of spectra obtained in the laboratory from well characterized samples. There is a rich literature on hyperspectral image classification (see~\cite{Survey} for a recent survey); however, classification methods emphasizing matching to a spectral library and material identification have received less attention~\cite{rivard2008continuous,prasad2012information,zhang2005wavelet}. On the one hand, many methods rely on nearest neighbor classification schemes based on one of many possible spectral similarity measures to match the observed test spectra with training library spectra. On the other hand, practitioners have designed feature extraction schemes that capture relevant information, in conjunction with appropriate similarity metrics, in order to discriminate between different materials.\par
Classification methods based on spectral similarity measures can provide researchers with simple implementation and relatively small computational requirements; however, there is a tradeoff with the amount of storage required for the training spectra as well as with the uneven performance of nearest neighbor methods. For example, in some cases taking the whole spectrum into consideration brings a large amount of redundant information to practitioners, while the role of relevant structural features is weakened. \par
Practitioners recognize several structural features in the spectral curves of each material as ``diagnostic" or characteristic of its chemical makeup, such as the position and shape of absorption bands. Several approaches like the Tetracorder \cite{RClark:imaging} have been proposed to encode such characteristics. However, such techniques require the construction of ad-hoc rules to characterize instances of each material while new rules must be created when spectral species which were not previously analyzed are added. Parente \emph{et al.} \cite{MParente:decomposition} proposed an approach using parametric models to represent the absorption features. However, it still requires the construction of specific rules to match observations to a training library.\par
In this paper, we consider the formulation of an information extraction process from hyperspectral signatures via the use of mathematical models for hyperspectral signals. Our goal is to encode the signature's scientifically meaningful structural features into numerical features, which are referred to as \emph{semantic} features, without ad-hoc rules for the spectra of any material type. Our proposed method provides automated extraction of semantic information from the hyperspectral signature, in contrast with the aforementioned diagnostic characteristics designed by hand by expert practitioners. Furthermore, no new rules should need to be constructed when mineral species which were not analyzed before are added. \par
Mathematical signal models have been used to represent reflectance spectra. More specifically, models leveraging wavelet decompositions are of particular interest because they enable the representation of structural features at different scales. The wavelet transform is a popular tool in many signal processing applications due to the capability of wavelet coefficients to characterize signal discontinuities at different scales and offsets. As mentioned above, the semantic information utilized by researchers is heavily related to the shape of reflectance spectra, which is succinctly represented in the magnitudes of its wavelet coefficients. A coefficient with large magnitude generally indicates a rapid change in its support while a small wavelet coefficient generally implies a smooth region. Existing wavelet approaches are limited to filtering techniques but do not extract features~\cite{rivard2008continuous,prasad2012information}.\par
In this paper, we apply hidden Markov models (HMMs) to the wavelet coefficients derived from the observed hyperspectral signals so that the correlations between wavelet coefficients at adjacent scales can be captured by the models. The HMMs allow us to identify significant vs.\ nonsignificant portions of the hyperspectral signatures with respect to the database used for training. The applications of HMMs for this purpose is inspired by the hidden Markov tree (HMT) model proposed in \cite{crouse1998wavelet}. As for the wavelet transform, we use an undecimated wavelet transform (UWT) in order to obtain maximum flexibility on the set of scales and offsets (spectral bands or wavelengths\footnotemark[1]) considered. \footnotetext[1]{We use these three equivalent terms interchangeably in the sequel.}\par
Our model for a spectrum encompassing $N$ spectral bands takes the form of a collection of $N$ non-homogeneous hidden Markov chains (NHMCs), each corresponding to a particular spectral band. Such a model provides a map from each signal spectrum to a binary space that encodes the structural features at different scales and wavelengths, effectively representing the semantic features that allow for the discrimination of spectra. To the best of our knowledge, the application of statistical wavelet models to the automatic selection of semantically meaningful features in hyperspectral signatures has not been proposed previously.\par
This paper is organized as follows. Section \ref{sec:bg} introduces the mathematical background behind our hyperspectral signature classification system and reviews relevant existing approaches for the hyperspectral classification task. Section \ref{sec:nhmc} provides an overview of the proposed feature extraction method, including details about the choice of mother wavelet, statistical model training, and label computing; we also show examples of the semantic information in hyperspectral signatures captured by the proposed features. Section \ref{sec:exps} describes our experimental test setup as well as the corresponding results. Some conclusions are provided in Section \ref{sec:conc}. Finally, proofs of our theoretical results are presented in the appendix.

\section{Background and Related Work}
\label{sec:bg}
In this section, we begin by discussing several existing spectral matching approaches. Then, we review the theoretical background for our proposed hyperspectral signature classification system, including wavelet analysis, hidden Markov chain models, and the Viterbi algorithm. 

\subsection{Spectral Matching Measures}
A direct comparison of spectral similarity measures taken on the observed hyperspectral signals is the easiest and the most direct way to do spectral matching. Generally speaking, spectral similarity measures can be combined with nearest neighbor classifiers. In this paper we use four commonly used spectral similarity measures. To present these measures, we use $\boldsymbol{r}_i=(r_{i1},r_{i2},...,r_{iN})^T$ and $\boldsymbol{r}_j=(r_{j1},r_{j2},...,r_{jN})^T$ to denote the reflectance or radiance signatures of two hyperspectral image pixel vectors
\subsubsection{Spectral Angle Measure}
The spectral angle measure (SAM)~\cite{FAKruse:the} between two reflectance spectra is defined as
\begin{equation*}
\mathrm{SAM}(\boldsymbol{r}_i,\boldsymbol{r}_j)=\cos^{-1}\left (\frac{\langle \boldsymbol{r}_i, \boldsymbol{r}_j\rangle}{\sqrt{||\boldsymbol{r}_i||_2^2||\boldsymbol{r}_j||_2^2}}\right ).
\end{equation*}
A smaller spectral angle indicates larger similarity between the spectra.
\subsubsection{Euclidean Distance Measure} 
The Euclidean distance measure (ED)~\cite{JNSweet:the} between two reflectance spectra is defined as
$ED(\boldsymbol{r}_i,\boldsymbol{r}_j)=||\boldsymbol{r}_i-\boldsymbol{r}_j||_2$.
As with SAM, smaller ED implies larger similarity between two vectors. The ED measure takes the intensity of two reflectance spectra into account, while the former is invariant to intensity.
\subsubsection{Spectral Correlation Measure} 
The spectral correlation measure (SCM)~\cite{FVDMeer:ccsm} between two reflectance spectra is defined as
\begin{equation*}
\mathrm{SCM}(\boldsymbol{r}_i,\boldsymbol{r}_j)=\frac{\sum_{k=1}^N (r_{ik}-\bar r_i)(r_{jk}-\bar r_j)}{\sqrt{\sum_{k=1}^N (r_{ik}-\bar r_i)^2 \sum_{k=1}^N (r_{jk}-\bar r_j)^2}}.
\end{equation*}
where $\bar r_i$ is the mean of the values of all the elements in a reflectance spectrum vector $\boldsymbol{r}_i$. The SCM can take both positive or negative values; larger positive values are indicative of similarity between spectra.
\subsubsection{Spectral Information Divergence Measure} 
The spectral information divergence measure (SID)~\cite{CIChang:an} between two reflectance spectra is defined as
$\mathrm{SID}(\boldsymbol{r}_i,\boldsymbol{r}_j)=D(\boldsymbol{r}_i||\boldsymbol{r}_j)+D(\boldsymbol{r}_j||\boldsymbol{r}_i)$,
where $D(\boldsymbol{r}_i||\boldsymbol{r}_j)$ is regarded as the relative entropy (or Kullback-Leibler divergence) of $\boldsymbol{r}_j$ with respect to $\boldsymbol{r}_i$, which is defined as
\begin{equation*}
D(\boldsymbol{r}_i||\boldsymbol{r}_j)=-\sum_{k=1}^N p_{ik}(\log p_{jk}-\log p_{ik}).
\end{equation*}
Here $p_{ik}=r_{ik}/\sum_{k=1}^N r_{ik}$ corresponds to a normalized version of the spectrum $r_i$ at the $k^\textrm{th}$ spectral band, which is interpreted in the relative entropy formulation as a probability distribution. \par

\subsection{Wavelet Analysis}
The wavelet transform of a signal provides a multiscale analysis of a signal's content which effectively encodes the locations and scales at which the signal structure is present in a compact fashion \cite{SMallat:a}. To date, several hyperspectral classification methods based on wavelet transform have been proposed. Most of these classification approaches (e.g. \cite{zhang2005wavelet, masood2008hyperspectral, west2007multiclassifiers}) employ a dyadic/decimated wavelet transform (DWT) as the preprocessing step. Compared with UWT, the DWT provides a more concise representation because it minimizes the amount of redundancy in the coefficients. However, the tradeoff for such redundancy is that UWT provides maximum flexibility on the choice of scales and offsets used in the multiscale analysis, which is desired because it allows for a simple characterization of the spectrum structure at each individual spectral band. \par
A one-dimensional real-valued UWT of an $N$-sample signal $x\in\mathbb{R}^N$ is composed of wavelet coefficients $w_{s}$, each labeled by a scale $l\in{1, ... , L}$ and offset $n\in{1, ... ,N}$, where $L\leqslant N$. The coefficients are defined using inner products as $w_{l,n}=\langle x,\phi _{l,n}\rangle$, where $\phi _{l,n} \in\mathbb{R}^N$ denotes a sampled version of the mother wavelet function $\phi$ dilated to scale $l$ and translated to offset $n$:
\begin{equation*}
\phi _{l,n}(\lambda)=\frac{1}{\sqrt{l}}\phi\left (\frac{\lambda -n}{l}\right ).
\end{equation*}
To improve the interpretability of the notation, we will change our notation for scales in the sequel from $l = 1,2,\ldots,L$ to $s = L,L-1,\ldots,1$ (i.e., we reverse the ordering of the scales). With this change, small values of $s$ correspond to coarse scales while large values of $s$ correspond to fine scales. All the coefficients can be organized into a two-dimensional matrix $W$ of size $L\times N$, where rows represent scales and columns represent wavelengths. In this case, each coefficient $w_{s,n}$, where $s<L$, has a child coefficient $w_{s+1,n}$ at scale $s+1$. Similarly, each coefficient $w_{s,n}$ at scale $s>1$ has one parent $w_{s-1,n}$ at scale $s-1$. Such a structure in the wavelet coefficients enables the representation of fluctuations in a spectral signature by chains of large coefficients appearing within the columns of the wavelet coefficient matrix $W$.\par

\subsection{Advantages of Haar Wavelet}
The Haar wavelet is the simplest possible compact wavelet which has the properties of square-like shape and discontinuity. These properties makes the Haar wavelet sensitive to a larger range of fluctuations than other mother wavelets and provides it with a lower discriminative power. Thus, the Haar wavelet enables the detection of both slow-varying fluctuations and sudden changes in a signal \cite{SMallat:a}, while not particularly sensitive to small discontinuities (i.e., noise) on a signal, in effect averaging them out over the wavelet support.\par
\begin{figure}[t]
\centering
\centerline{\includegraphics[width=9cm]{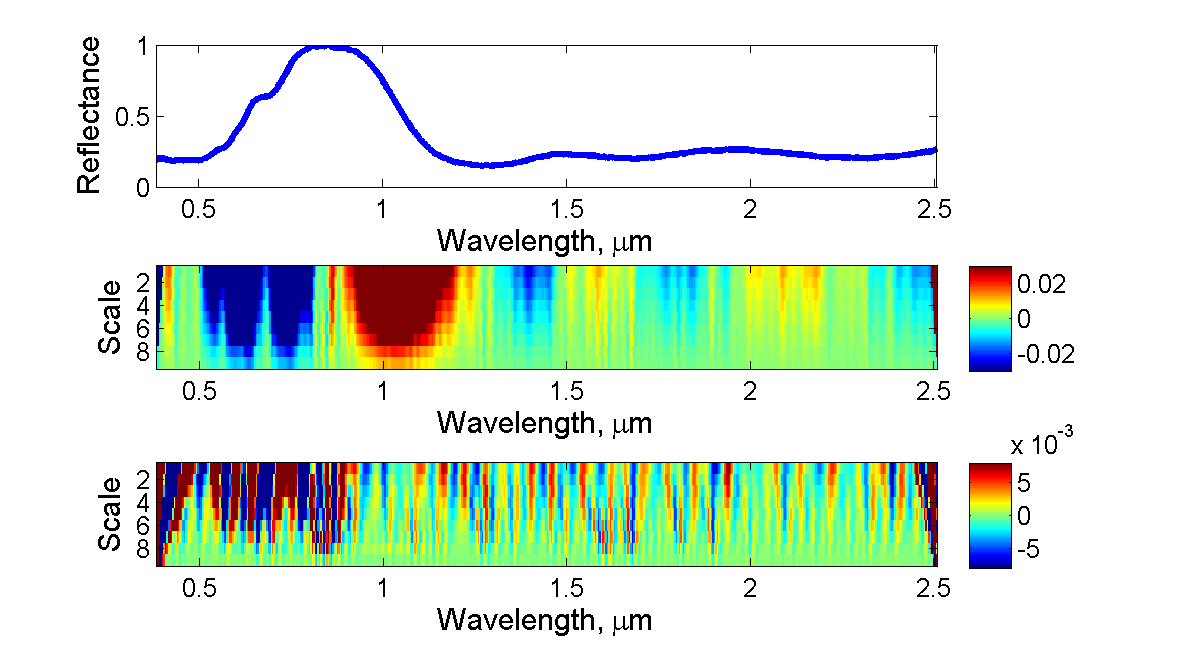}}
\caption{Top: an example of normalized mineral reflectance (Garnet). Middle: corresponding UWT coefficient matrix (9-level wavelet decomposition) using a Haar wavelet. Bottom: corresponding UWT coefficient matrix using a Daubechies-4 wavelet.}
\label{fig:two}
\end{figure}

Consider the example in Fig.~\ref{fig:two}, where the figure at the top represents an example hyperspectral signature, while the figures in the middle and at the bottom show the undecimated wavelet coefficient matrix of the spectrum under the Haar and Daubechies-4 wavelets, respectively. The middle figure in Fig.~\ref{fig:two} shows the capability of Haar wavelets to capture both rapid changes and gently sloping fluctuations in the sample reflectance spectrum.
Similarly, the bottom figure shows that the Daubechies-4 wavelet is sensitive to compact and drastic discontinuities (i.e., higher order fluctuations that are often due to noise). 
Thus, the Daubechies-4 wavelet does not provide a good match to semantic information extraction for this example reflectance spectrum. Intuitively, these issues will also be present for other higher-order wavelets, which provide good analytical matches to functions with fast, high-order fluctuations.\par
In general, wavelet representations of spectral absorption bands are less emphasized under Haar wavelet than under other higher order wavelets. However, this drawback can be alleviated using discretization, which will be described in the next subsection. 

\subsection{Statistical Modeling of Wavelet Coefficients}
\label{sec:bgnhmc}
Crouse \emph{et al.} \cite{crouse1998wavelet} proposed the use of hidden Markov models (HMM) to capture the statistics of DWT coefficients. In that paper, the dyadic nature of DWT coefficients gives rise to 
a hidden Markov tree (HMT) model that characterizes the clustering and persistence properties of wavelet coefficients. The statistical model is constructed based on the wavelet representation of spectra in a training library. 

The statistical model is motivated by the compression property of the DWT, which states that the wavelet transform of a piecewise smooth signal generally features a small number of large coefficients and a large number of small coefficients. This property motivates the use of a zero-mean Gaussian mixture model (GMM) with two Gaussian components to capture the compression property, where one Gaussian component with a high-variance characterizes the small number of ``large" coefficients (labeled with a state $\mathrm{L}$), while a second Gaussian component with a low-variance characterizes the large number of ``small" wavelet coefficients (labeled with a state $\mathrm{S}$). The state $S_s\in \{\mathrm{S},\mathrm{L}\}$ of a wavelet 
coefficient\footnote{Since the same model is used for each chain of coefficients $\{S_{1,n},\ldots,S_{L,n}\}$, $n=1,\ldots,N$, we remove the index $n$ from the subscript for simplicity in this sequel whenever possible.} 
is said to be hidden because its value is not explicitly observed. The likelihoods of the two Gaussian components $p_{S_s}(\mathrm{L}) = p(S_s = \mathrm{L})$ and $p_{S_s}(\mathrm{S}) = p(S_s = \mathrm{S})$ should meet the condition that $p_{S_s}(\mathrm{L})+p_{S_s}(\mathrm{S})=1$. The conditional probability of a particular wavelet coefficient $w_{s}$ given the value of the state $S_{s}$ can be written as $p(w_{s}|S_{s}=i)=\mathcal{N}(0,\sigma_{i,s}^2)$, where $i=\{ \mathrm{S}, \mathrm{L}\}$, and the distribution of the same wavelet coefficient can be written as $p(w_s)=p_{S_s}(\mathrm{L})\mathcal{N}(0,\sigma_{\mathrm{L},s}^2)+p_{S_s}(\mathrm{S})\mathcal{N}(0,\sigma_{\mathrm{S},s}^2)$.\par
In cases where a UWT is used, the persistence property of wavelet coefficients \cite{SMallat:singularity, SMallat:characterization} (which implies the high probability of a chain of wavelet coefficients to be consistently small or large across adjacent scales) can be accurately modeled by a non-homogeneous hidden Markov chain (NHMC) that links the states of wavelet coefficients in the same offset. This means the state $S_{s}$ of a coefficient $w_{s}$ is only affected by the state $S_{s-1}$ of its parent (if it exists) and by the value of its coefficient $w_s$. The Markov chain is completely determined by the likelihoods for the first state and the set of state transition matrices for the different parent-child label pairs $(S_{s-1},S_{s})$ for $s>1$:
\setcounter{MaxMatrixCols}{100}
\begin{equation}
A_{s}=
\begin{pmatrix}
p_{\mathrm{S} \rightarrow \mathrm{S},s} & p_{\mathrm{L} \rightarrow \mathrm{S},s}\\
p_{\mathrm{S} \rightarrow \mathrm{L},s} & p_{\mathrm{L} \rightarrow \mathrm{L},s}
\end{pmatrix}
,
\label{eq:tmat}
\end{equation}
where $p_{i\to j,s} := P(S_s = j | S_{s-1}=i)$ for $i,j \in \{\mathrm{L},\mathrm{S}\}$.
The training process of an HMM is based on the expectation maximization (EM) algorithm which generates a set of HMM parameters  \\
$\boldsymbol{\theta} = \{p_{S_1}(\mathrm{S}),p_{S_1}(\mathrm{L}),\{A_s\}_{s=2}^L,\{\sigma_{\mathrm{S},s},\sigma_{\mathrm{L},s}\}_{s=1}^L\}$ including the probabilities for the first hidden states, the state transition matrices, and Gaussian variances for each of the states. 
We define the $L\times N$ matrix $S$ containing the collection of state values for all scales and spectral bands. 
The iterative parts of the algorithm can be briefly described as follows:
\begin{enumerate}
\item \textbf{E step}: Perform maximum likelihood estimation of the state labels using a forward-backward algorithm \cite{rabiner1989tutorial}: 
$$S^l = \arg \max_{S} p(S|W, \boldsymbol{\theta}^l);$$ 
this joint conditional probability mass function (PMF) will be used in the M step.
\item \textbf{M step}: Update model parameters to maximize the expected value of the joint likelihood of the wavelet coefficients and state estimates~\cite{crouse1998wavelet}: 
$$\boldsymbol{\theta}^{l+1}=\arg\min_{\boldsymbol{\theta}} E_{S}[\ln f(W,S|\boldsymbol{\theta}^l)|W,\boldsymbol{\theta}^l].$$ 
\item Set $l=l+1$. If converged, then stop; otherwise, repeat.
\end{enumerate}
\subsection{Wavelet-based Spectral Matching}
\label{sec:wavspec}
Many hyperspectral signature classification approaches have been proposed in the literature, with a subset of them involving wavelet analysis \cite{rivard2008continuous,prasad2012information,zhang2005wavelet,qian2013hyperspectral,shen2011three}. In this paper, we review two approaches that are particularly close in scope to our proposed method, which will be used for comparison in our numerical experiments. Since our focus in this paper is on hyperspectral classification for individual pixels, we limit our comparison to methods that rely exclusively on the spectral of a given pixel or on features obtained from the pixel's spectra. More specifically, we do not compare to other methods that use additional information (e.g. spatial information for a HSI) or that assume prior knowledge of the location of semantic information, which is usually obtained from an expert practitioner. \par
First, Rivard et al.~\cite{rivard2008continuous} propose a method based on the wavelet decomposition of the spectral data. The obtained wavelet coefficients are separated into two categories: low-scale components of power (LCP) capturing mineral spectral features (corresponding to the first fine scales), and high-scale components of power (HCP) containing the overall continuum (corresponding to coarser scales). The coefficients for the LCP spectrum, which capture detailed structural features, are summed across scales at each spectral band. This process can conceptually be described as a filtering approach, since the division into LCP and HCP effectively acts as a high-pass filter that preserves only the fine-scale detailed portion of the spectrum.\par
A second wavelet-based classification approach is proposed in \cite{prasad2012information}. This second approach applies an UWT on the entire database. The set of wavelet coefficients for each separate wavelength is considered as a separate feature vector. Linear discriminant analysis (LDA) is performed on each one of these vectors for dimensionality reduction purposes. The outputs are grouped into $C$ classes, corresponding to the elements of interest, to train either a single multivariate Gaussian distribution or a GMM for each of the classes, where a classification label or score is obtained for each wavelength. Finally, decision fusion is performed among the wavelengths to obtain a single classification label for the spectrum.
It is implicitly expected by this method that the number of training samples for each one of the classes is sufficiently large so that the class-specific Gaussian (mixture) models can be accurately constructed.\par

\section{NHMC-Based Feature Extraction and Classification}
\label{sec:nhmc}

In this section, we introduce a feature extraction scheme for hyperspectral signatures that exploits a Markov model for the signature's wavelet coefficients. A wavelet analysis is used in an UWT to capture information on the fluctuations of the spectra. The state labels extracted from the Markov model represent the semantic information relevant for hyperspectral signal processing. 

\subsection{Multi-State Hidden Markov Chain Model}
In our system, we choose to use the NHMC model described in Section~\ref{sec:bgnhmc} applied to the UWT via the Haar wavelet. We select the Haar wavelet due to its special shape, which allows for the magnitude of the wavelet coefficients to be proportional to the slope of the spectra across the wavelet's support. Furthermore, the signs of these coefficients are indicative of the slope orientation (increasing or decreasing for negative and positive, respectively).\par
In contrast to the prior work of~\cite{crouse1998wavelet}, we design our NHMC to feature $k$-state GMMs for the wavelet coefficients. We increase the number of states from 2 to $k>2$ because a two-state zero-mean GMM provides an overly coarse distinction between sharper absorption bands (fluctuations) and flatter regions in a hyperspectral signature, which are usually assigned large and small state labels, respectively. In our cases of interest, spectrum classification requires a labeling granularity for the signature fluctuations that is finer than that achieved by binary labels. \par
We associate each wavelet coefficient $w_{s}$ with an unobserved hidden state $S_{s}\in \{0,1,...,k-1\}$, where the states have prior probabilities $p_{i,s}:=p(S_{s}=i)$ for $i=0,1,...,k-1$. Here the state $i=0$ represents smooth regions of the spectral signature, in a fashion similar to the small ($\mathrm{S}$) state for binary GMMs, while $i = 1,\ldots,k-1$ represent a more finely grained set of states for spectral signature fluctuations, similarly to the large ($\mathrm{L}$) state in binary GMMs. All the weights should meet the condition $\sum_{i=0}^{k-1} p_{i,s}=1$. Each state is characterized by a zero-mean Gaussian distribution for the wavelet coefficient with variance $\sigma_{i,s}^2$. The value of $S_{s}$ determines which of the $k$ components of the mixture model is used to generate the probability distribution for the wavelet coefficient $w_{s}$: $p(w_{s}|S_{s}=i)=\mathcal{N}(0,\sigma_{i,s}^2)$. We can then infer that $p(w_{s})=\sum_{i=0}^{k-1} p_{i,s}p(w_{s}|S_{s}=i)$. In analogy with the binary GMM case, we can also define a $k\times k$ transition probability matrix
\setcounter{MaxMatrixCols}{100}
\begin{equation*}
A_{s}=
\begin{pmatrix}
p_{0\rightarrow 0,s} & p_{1\rightarrow 0,s} & \cdots & p_{k-1\rightarrow 0,s}\\
p_{0\rightarrow 1,s} & p_{1\rightarrow 1,s} & \cdots & p_{k-1\rightarrow 1,s}\\
\vdots & \vdots & \ddots & \vdots\\
p_{0\rightarrow k-1,s} & p_{1\rightarrow k-1,s} & \cdots & p_{k-1\rightarrow k-1,s}
\end{pmatrix}
,
\end{equation*}
where $p_{i\rightarrow j,s}=p(S_{s}=j|S_{s-1}=i)$. Note that the probabilities in the diagonal of $A_{s}$ are expected to be larger than those in the off-diagonal elements due to the persistence property of wavelet transforms. Note also that all state probabilities ${p_{i,s}}$ for $s > 1$ can be derived from the matrices $\{A_{s}\}_{s=2}^L$ and $\{p_{i,1}\}_{i=0}^{k-1}$.\par
The training of the $k$-GMM NHMC is also performed via an EM algorithm. Because of the overlap between wavelet functions at a fixed scale and neighboring offsets, adjacent coefficients may have correlations in relative magnitudes \cite{MTOrchard:an}. However, for computational reasons, in this paper we only consider the parent-child relationship of the wavelet coefficients in the same offset. Namely, we train an NHMC separately on each of the $N$ wavelengths sampled by the hyperspectral acquisition device. The set of NHMC parameters $\boldsymbol{\theta}_n$ of a certain spectral band $n$ include the probabilities for the first hidden states $\{p_{i,1,n}\}_{i=0}^{k-1}$, the state transition matrices $\{A_{s,n}\}_{s=2}^L$, and the Gaussian variances $\{\sigma_{0,s,n}^2,\sigma_{1,s,n}^2,\ldots,\sigma_{k-1,s,n}^2\}_{s=1}^L$. In the sequel, we remove from the parameters $\boldsymbol{\theta}$ the dependence on the wavelength index $n$ whenever possible.

\subsection{Label Computation}
\label{sec:label}
Given the model parameters $\boldsymbol{\theta}$, the state label values $\{S_{s}\}_{s=1}^L$ for a given observation are obtained using a Viterbi algorithm~\cite{rabiner1989tutorial,crouse1998wavelet}. For a particular wavelet coefficient $w_{s}$, a $k$-dimensional conditional probability vector is defined with elements being the conditional PMF of the wavelet coefficient $$p(w_{s}|S_{s}=i)=\frac{1}{\sqrt{2\pi \sigma_{s}^2}}\exp \left(-\frac{{w_{s}}^2}{2\sigma_{s}^2}\right)$$ under each possible state value $i=0,\ldots,k-1$. A variable $\delta_{i,s}$ is defined as the ``best score" that ends in a particular state $i$ at scale $s$ from its previous state, while the variable $\psi_{i,s}$ is the most likely state at a particular scale $s-1$ to have children $s$ with state $i$. The definitions of the two variables are
\begin{align}
\psi_{i,1}&=0,\label{eq:viterbi1}\\
\delta_{i,1}&=p_{i,1}\cdot p(w_{1}|S_{1}=i),\\
\psi_{i,s}&=\arg\max_{j=0,\ldots,k-1}(\delta_{j,s-1}p_{j \rightarrow i,s}),\\
\delta_{i,s}&=\delta_{\psi_{i,s},s-1}p_{\psi_{i,s} \rightarrow i,s}\cdot p(w_{s}|S_{s}=i),\label{eq:viterbi4}
\end{align}
for $i=1,\ldots,k-1$ and $s = 2,\ldots,L$. The algorithm also returns the likelihood $p(W|\boldsymbol{\theta})$ of a wavelet coefficient matrix $W$ under the model $\boldsymbol{\theta}$ as a byproduct. We propose the use of the state label array $S$ as classification features for the original hyperspectral signal $x$. It is easy to identify the presence of such features simply by inspecting the labels obtained from the NHMC. 

\subsection{Additional Modifications to NHMC}
\begin{figure}[t]
\centering
\centerline{\includegraphics[width=9cm]{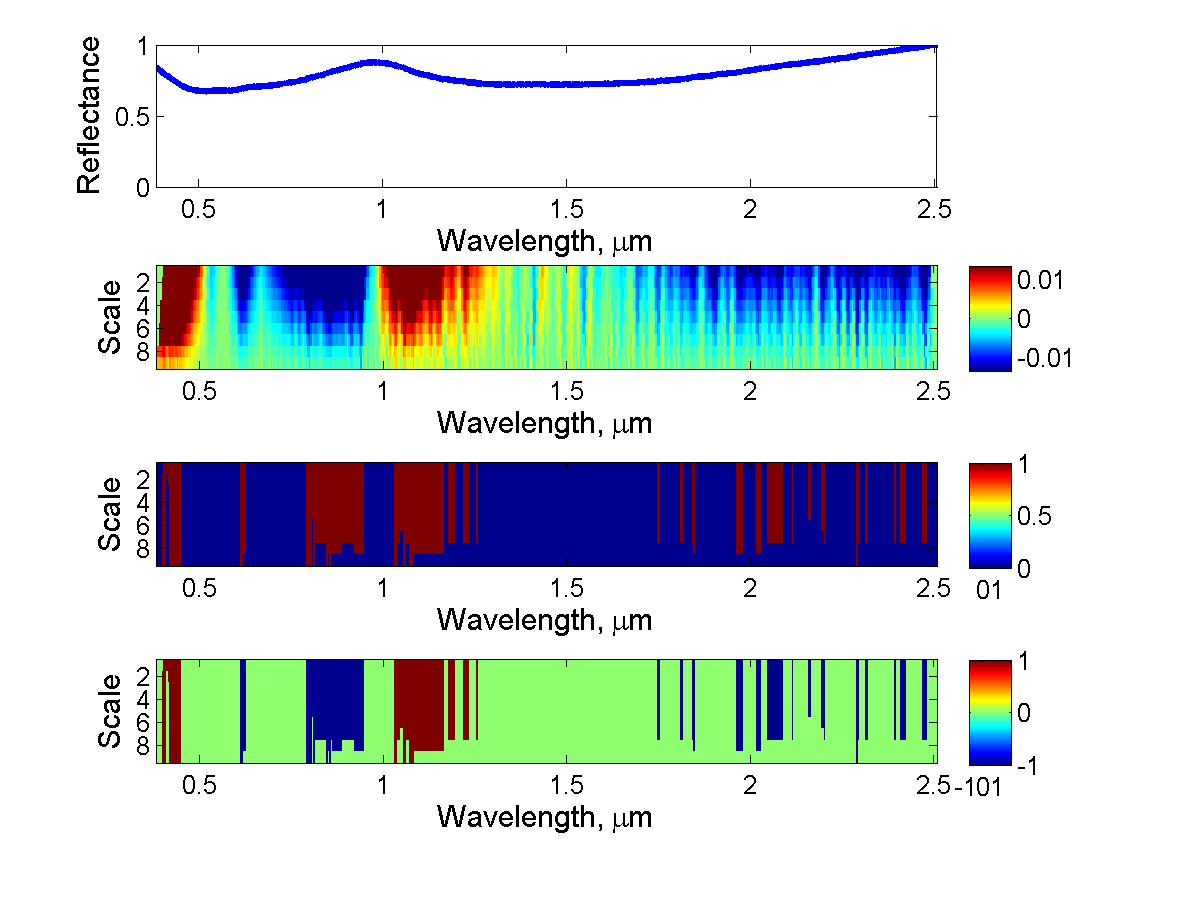}}
\vspace{-5mm}
\caption{Examples of signed state labels as classification features. Top: Example normalized reflectance spectrum (Ilmenite). Second: Corresponding 9-label UWT coefficient matrix using a Haar wavelet. Third: Corresponding state label matrix from an NHMC model using a zero-mean two-state GMM. Blue represents smooth regions, while red represents fluctuations. Bottom: Corresponding features consisting of the state labels with added signs from the Haar wavelet coefficients. Green represents smooth regions, while red represents decreasing fluctuations and blue represents increasing fluctuations.}
\label{fig:three}
\end{figure}

As mentioned above, because of the shape of the Haar wavelet function, the signs of Haar wavelet coefficients of a reflectance spectrum capture whether the slopes increase or decrease as a function of wavelength. This characteristic of Haar wavelet coefficients can be utilized to design state labels that capture the slope orientations of the corresponding reflectance spectra. Thus, we make a simple modification by adding the sign of a Haar wavelet coefficient to its counterpart in the corresponding state label matrix. 
Fig.~\ref{fig:three} shows the effect of adding signs to state label matrices. The top two figures represent the reflectance spectrum of a sample material and the corresponding Haar wavelet coefficient matrix, while the bottom two show the corresponding state label matrices with and without being added wavelet coefficient signs, respectively. 
The figure shows that the fluctuations in the region $0.6-0.8~\textrm{\mum}$ are predominantly not detected by state labels. Furthermore, one can see many narrow chains of ``large'' state labels starting at $1.7~\textrm{\mum}$. Increasing the number of GMM state enables a finer-scale quantization of spectral signature fluctuations, which is somewhat analogous to increasing the quantization resolution for our wavelet analysis. This is quite important when the Haar wavelet is used due to its sensitivity to a large range of fluctuation orders, which implies a relatively low discriminative power when compared with higher-order wavelet transforms.\par
Unfortunately, a large number of GMM states might also have negative influence on classification results. The GMM state of a particular wavelet coefficient $w_{s,n}$ is determined by the coefficient's magnitude with respect to those for the rest of the NHMC training spectra, the state label of its parent $S_{s-1,n}$, and the transition probability matrix $A_{s,n}$. In practice, this dependence causes different maps between coefficient value ranges and GMM states across scales and offsets $(s,n)$. This variance often makes it difficult to assess the semantic information in the label array of a spectral signature. In practice, this variance may sometimes affect the interpretability of features obtained from GMM labels. Furthermore, the likelihood of such variability in the value-to-state mappings could increase when we use multi-state GMM. Additionally, such variance may have a particularly negative influence on classification schemes based on NN classifiers that act on GMM state label vectors. Thus, we desire a modification to the model that features the simplicity of a binary-state GMM (to preclude mismatch in coefficient-to-state mappings across wavelengths and states) and the spectral fluctuation characterization capability of a multi-state GMM (providing finer fluctuation characterization than a binary-state GMM).\par
We propose a solution that combines the advantages of a binary-state GMM and a $k$-state GMM, where $k>2$. Our modified wavelet coefficient statistical model consists of a binary-state NHMC with a ``small'' state ($0$) modeled by a standard zero-mean Gaussian distribution and a ``large'' state ($1$) modeled by a mixture of $k$-1 Gaussian distributions. Note that we use numbers here instead of letters for the state labels to distinguish between the $2$-state GMM NHMC and the $2$-state MOG NHMC. We denote this modified model mixture of Gaussians (MOG) NHMC in the sequel. As desired, this modified model maintains the discriminability between smooth regions and absorption bands in spectral signatures, while providing classification features (binary labels, in this case) that decrease the likelihood of the variability stated above.\par
In order to obtain a MOG NHMC model, the first step is to train a $k$-state GMM NHMC model that yields state labels $S_s \in \{0,\ldots,k-1\}$. After that, all the states are quantized into two states so that we can get a MOG NHMC that yields state labels $Z_s \in \{0,1\}$ with probabilities $q_{i,s} = P(Z_s = i)$, $i=0,1$. 
One can show that the change of models lead to the following mapping for labels:
\begin{equation}
Z(S) = \left\{\begin{array}{ll} 0 & \textrm{if } S = 0,\\1 & \textrm{if } S \ne 0.\end{array}\right.
\label{eq:map}
\end{equation}
Similar to~(\ref{eq:tmat}), we can define a transition probability matrix 
\begin{equation*}
B_{s}=
\begin{pmatrix}
q_{0 \rightarrow 0,s} & q_{1 \rightarrow 0,s}\\
q_{0 \rightarrow 1,s} & q_{1 \rightarrow 1,s}
\end{pmatrix}
\end{equation*}
for the MOG NHMC, where $q_{i\to j,s} := P(Z_s = j | Z_{s-1}=i)$ for $i,j \in \{0,1\}$ and $s=1,\ldots,L$. We have the following pair of intuitive results, whose proves are presented in Appendices~\ref{app:lemmaone} and~\ref{app:mog}.
\begin{lemma}
Denote the vector of state probabilities for a wavelet coefficient $w_{s}$ under the $k$-state GMM NHMC as $\boldsymbol{P}_{s}=(p_{0,s}, p_{1,s}, ..., p_{k-1,s})^T$. The corresponding vector of probabilities for the MOG NHMC states $\boldsymbol{Q}_{s}$ can be written as follows:
\begin{equation*}
\boldsymbol{Q}_{s}=(q_{0,s},q_{1,s})^T=\left(p_{0,s}, \sum_{i=1}^{k-1} p_{i,s}\right)^T = \left(p_{0,s}, 1 -p_{0,s}\right)^T.
\end{equation*}
\label{lemma:one}
\end{lemma}
\begin{lemma}
The elements of the MOG NHMC transition matrix $\boldsymbol{B}_s$ can be written in terms of the elements of the GMM NHMC transition matrix $\boldsymbol{A}_s$ as follows:
\begin{align}
q_{0 \rightarrow 0, s}&=p_{0 \rightarrow 0, s},\label{eq:lem1}\\
q_{1 \rightarrow 0,s}&=\frac{\sum_{i=1}^{k-1}p_{i \rightarrow 0,s}p_{i,s-1}}{\sum_{i=1}^{k-1} p_{i,s-1}},\\
q_{0 \rightarrow 1,s}&=\sum_{j=1}^{k-1}p_{0 \rightarrow j,s},\\
q_{1 \rightarrow 1,s}&=\frac{\sum_{i=1}^{k-1} p_{i,s-1}\sum_{j=1}^{k-1} p_{i \rightarrow j,s}}{\sum_{i=1}^{k-1} p_{i,s-1}}.\label{eq:lem4}
\end{align}
Here $i$ and $j$ represent state labels ranging from $1$ to $k-1$.
\label{lem:mog}
\end{lemma}
Below is an example of the transform of a state probability vector and transition probability matrix, respectively, where the original number of state is 4:
\begin{equation*}
(0.422, 0.3696, 0.1042, 0.1042)^T \rightarrow (0.422, 0.578)^T,
\end{equation*}
\begin{equation*}
\begin{pmatrix}
1 & 0.0001 & 0 & 0 \\
0 & 0.9999 & 0 & 0 \\
0 & 0 & 0.5 & 0.4999 \\
0 & 0 & 0.5 & 0.5001
\end{pmatrix}
\rightarrow
\begin{pmatrix}
1 & 0 \\
0 & 1
\end{pmatrix}.
\end{equation*}
\par
\begin{figure}[t]
\centering
\centerline{\includegraphics[width=9cm]{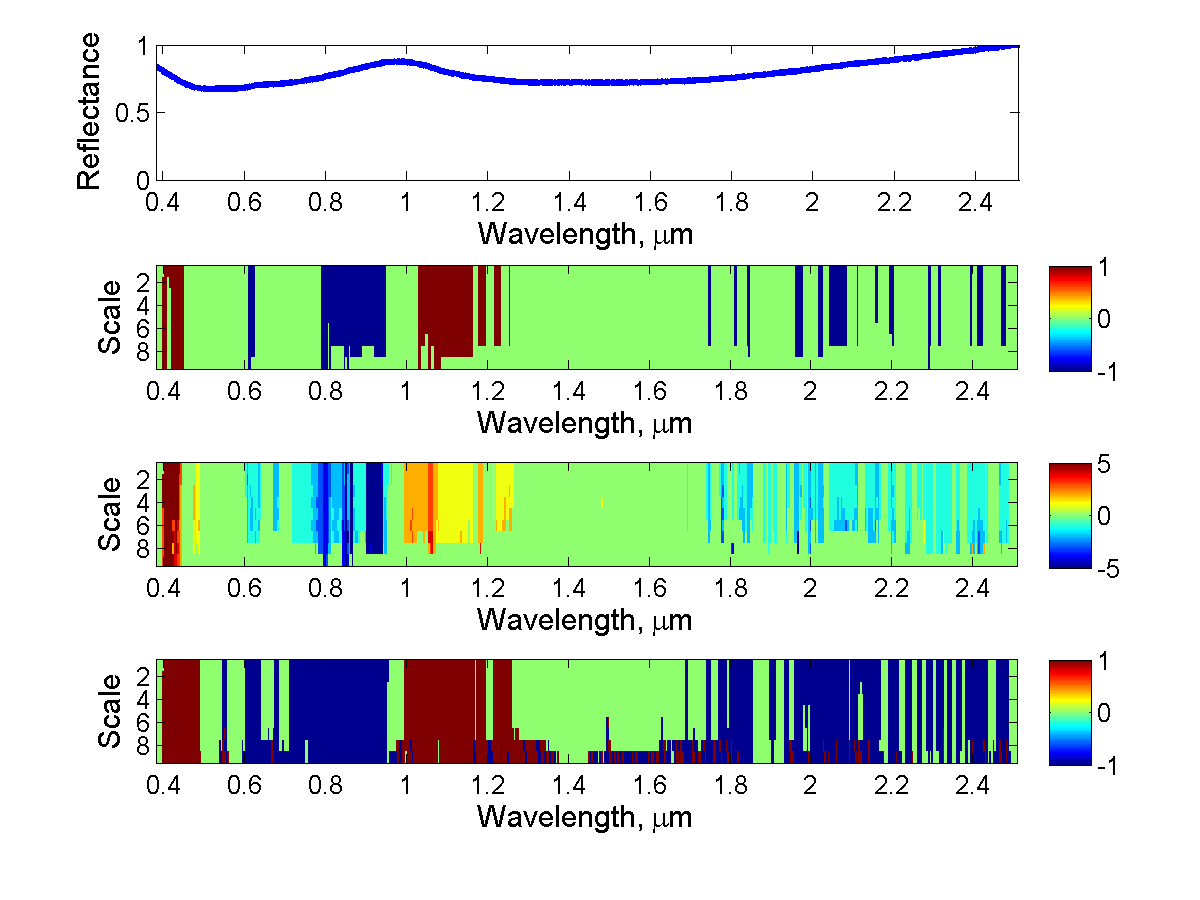}}
\caption{Comparison of label arrays obtained from several statistical models for wavelet coefficients. Top: example normalized reflectance, same as in Fig. \ref{fig:three}. Second: Corresponding state label matrix from a 2-state GMM NHMC model. Third: Corresponding state label matrix from a 6-state GMM NHMC model. Bottom: Corresponding state label matrix from a MOG NHMC model with $k=6$.}
\label{fig:four}
\end{figure}

Correspondingly, we also make small modifications to the label computation scheme from Section~\ref{sec:label}. For the MOG NHMC, equations (\ref{eq:viterbi1}--\ref{eq:viterbi4}) become
\begin{align*}
\psi_{i,1}&=0,\\
\delta_{i,1}&=q_{i,1}\cdot p(w_{1}|Z_{1}=i),\\
\psi_{i,s}&=\arg\max_{j=0,1}(\delta_{j,s-1}q_{j \rightarrow i,s}),\\
\delta_{i,s}&=\delta_{\psi_{i,s},s-1}q_{\psi_{i,s} \rightarrow i,s}\cdot p(w_{s}|Z_{s}=i),\end{align*}
respectively, for $i=0,1$ and $s=2,\ldots,L$. 
The required conditional probabilities involving $Z_s$ can be written as given in the following lemma.
\begin{lemma}
The state-conditional probabilities for the MOG NHMC can be given in terms of the state-conditional probabilities for the GMM NHMC as follows:
\begin{align*}
p(w_{s}|Z_{s}=0)&=p(w_{s}|S_{s}=0),\\
p(w_{s}|Z_{s}=1)&=\frac{\sum_{i=1}^{k-1} p_{i,s}p(w_{s}|S_{s}=i)}{\sum_{i=1}^{k-1} p_{i,s}},
\end{align*}
where $i$ denotes a state label ranging from $1$ to $k-1$. 
\end{lemma}
We provide an example comparison between labels obtained from the GMM NHMC and the MOG NHMC in Fig.~\ref{fig:four}. \par

\subsection{Illustration of Extracted Semantic Information}
\begin{figure*}[t]
\centering
\centerline{
\includegraphics[width=6cm]{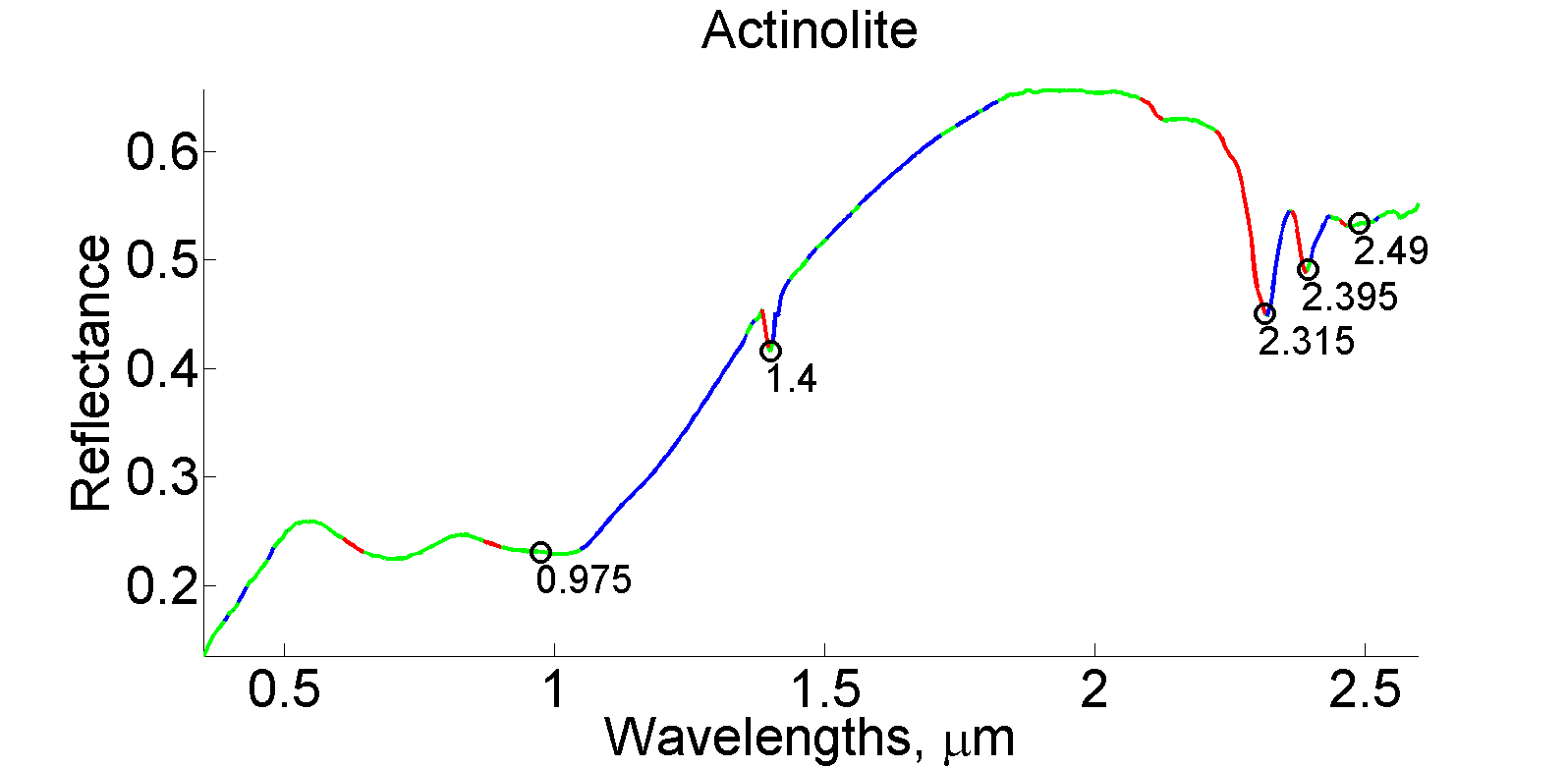}
\includegraphics[width=6cm]{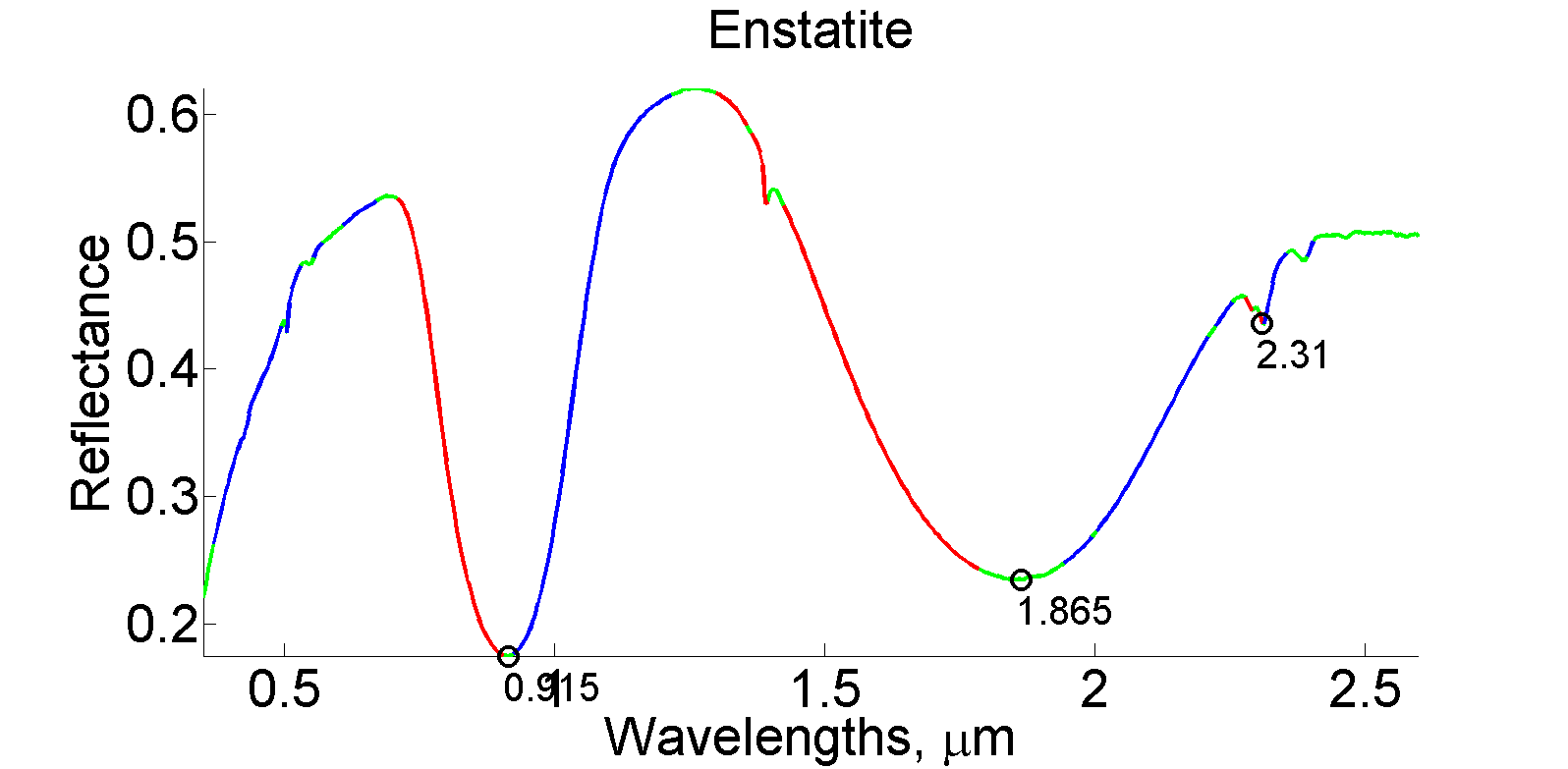}
\includegraphics[width=6cm]{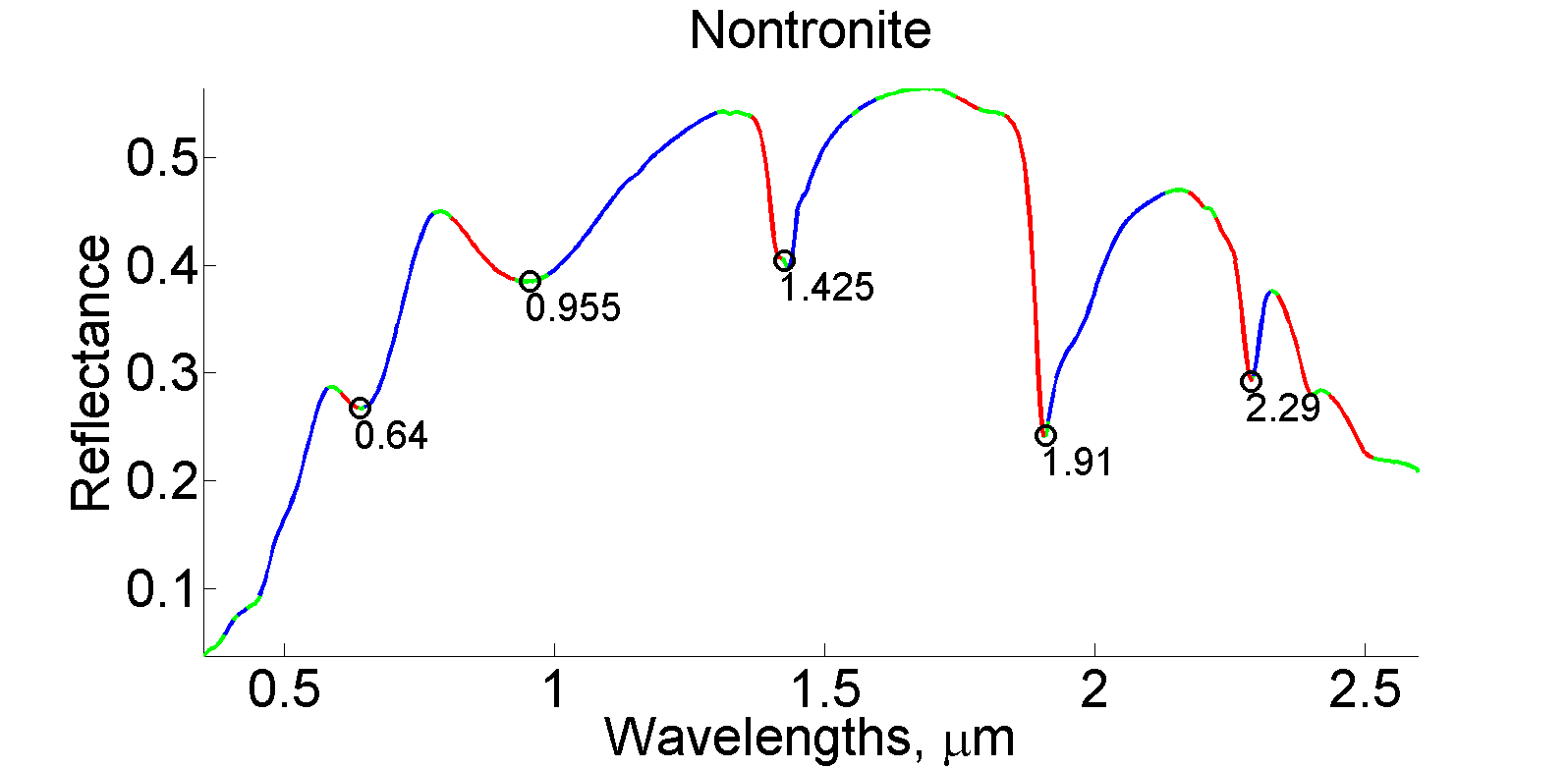}}
\vspace{0.2cm}
\centering
\centerline{
\includegraphics[width=6cm]{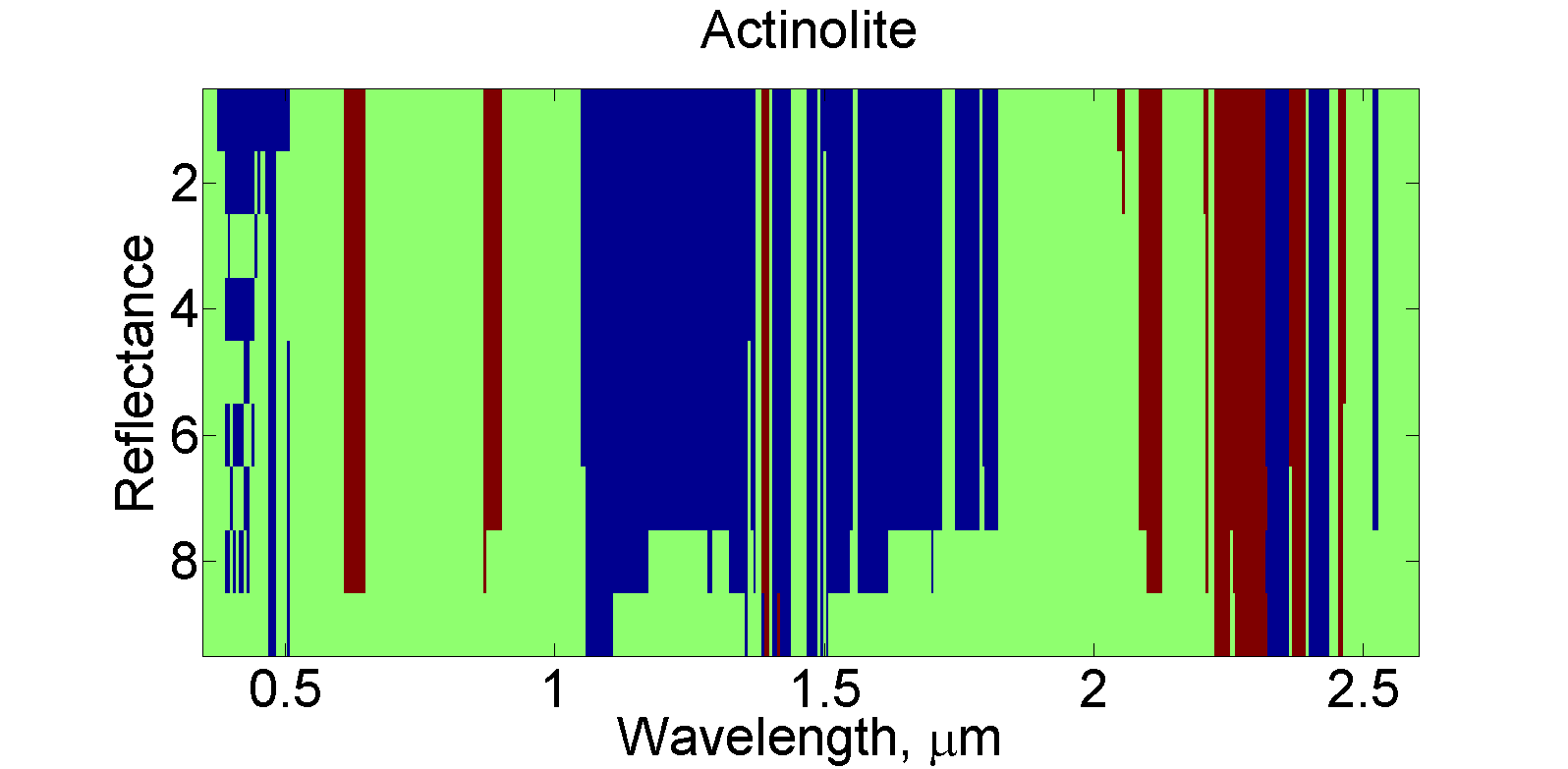}
\includegraphics[width=6cm]{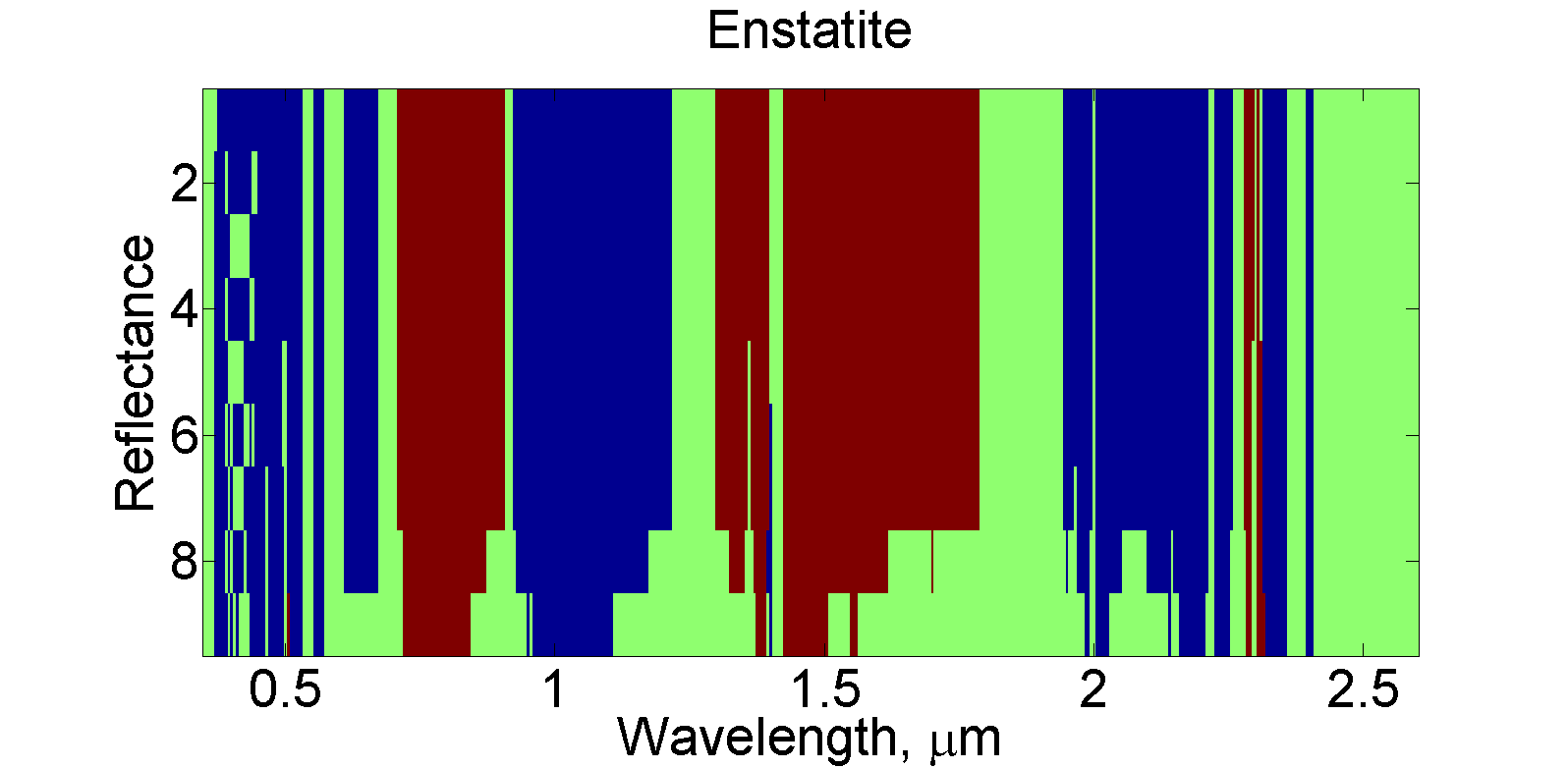}
\includegraphics[width=6cm]{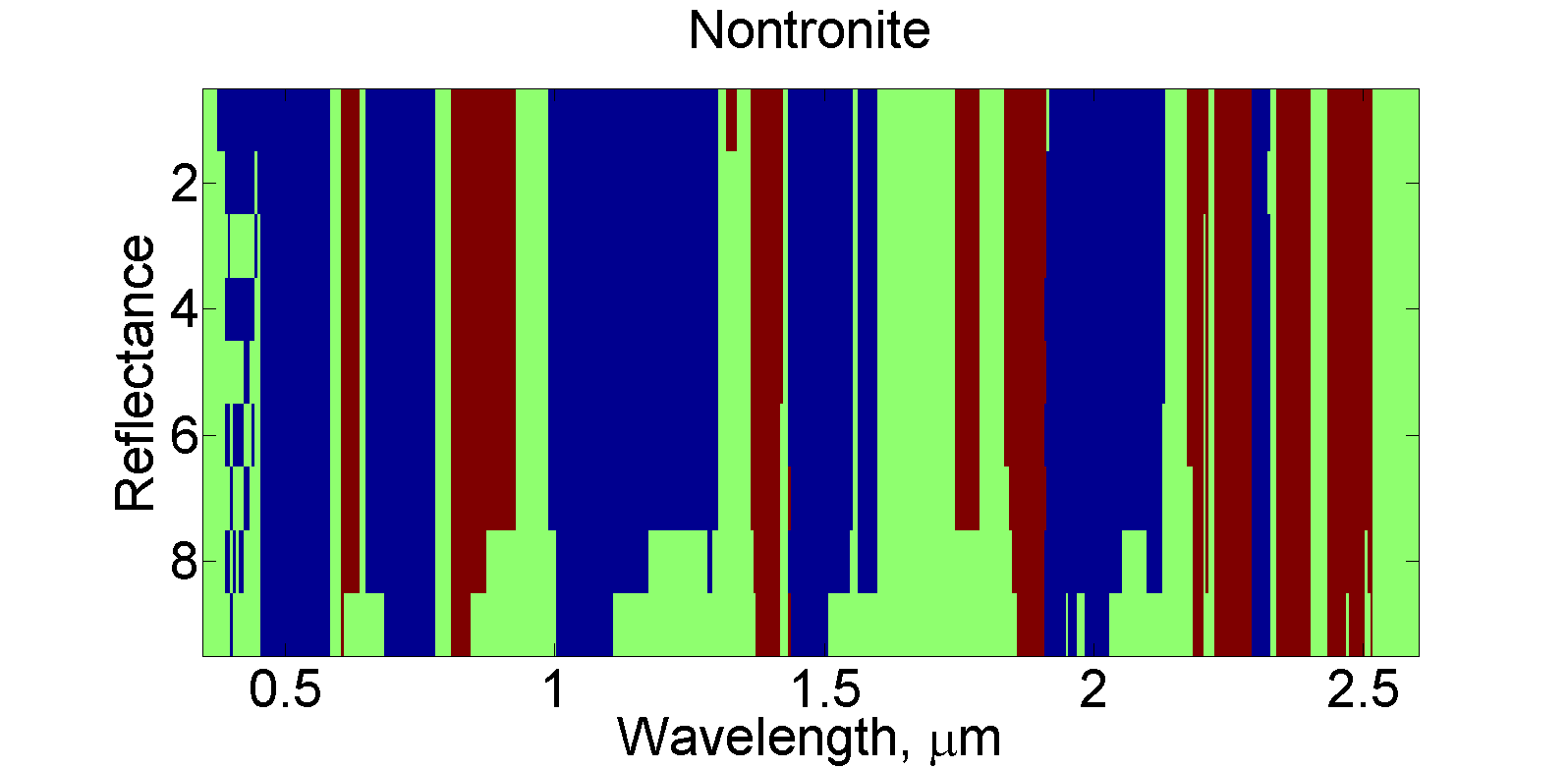}}
\caption{Semantic information extracted in some sample spectral curves based on MOG with 2 states. Top row: Sample spectral curves with extracted semantic information. Bottom row: corresponding state label array. }
\label{fig:semantic}
\end{figure*}

The state label arrays obtained from the NHMC model characterize four important semantic features of the corresponding hyperspectral signatures: ($i$) the orientations of the signature slope, which is reflected in the state label values; ($ii$) the extent of the signature slope, which is reflected in the duration of corresponding state label values through different wavelengths; ($iii$) the intensity of the signature slope, which is reflected on the depth of the corresponding state label values through the scales (when GMMs are used); and ($iv$) the locations of the absorption bands, which are reflected in the locations at which the labels switch from $+1$ to $-1$. In order to showcase the semantic information captured by our designed features, we illustrate these four types of semantic features in several example reflectance spectra. For convenience of illustration, we only use state label arrays based on MOG due to its binary property, which only reflects the orientation of slopes regardless of the intensities. To begin, we calculate the mean of each column in a state label array and then transform it to an integer by using round. In this way, we obtain what we call a state label mean vector of the same length as the corresponding reflectance spectrum whose possible element values are $\pm 1$ and $0$. Figure~\ref{fig:semantic} shows four example reflectance spectra with the corresponding extracted semantic information based on an NHMC using an MOG with 2 states as well as the corresponding state label arrays. We plot the reflectance spectral curve by using three different colors to encode the value of the state label mean vector: green, red, and blue portions represent wavelengths for which the state label mean vector elements are $0$, $+1$, and $-1$, respectively. Finally, we calculate all the middle points between the end of a $1$'s series and the beginning of a $-1$'s series, and mark those points on the plotted reflectance spectra to find the locations of absorption bands.
As expected, spectral curves in Fig.~\ref{fig:semantic} have blue increasing slopes, red decreasing slopes, and green flat regions.\par

\subsection{Classification System Overview}
\begin{figure*}[t!]
\centering
\centerline{\includegraphics[width=17cm]{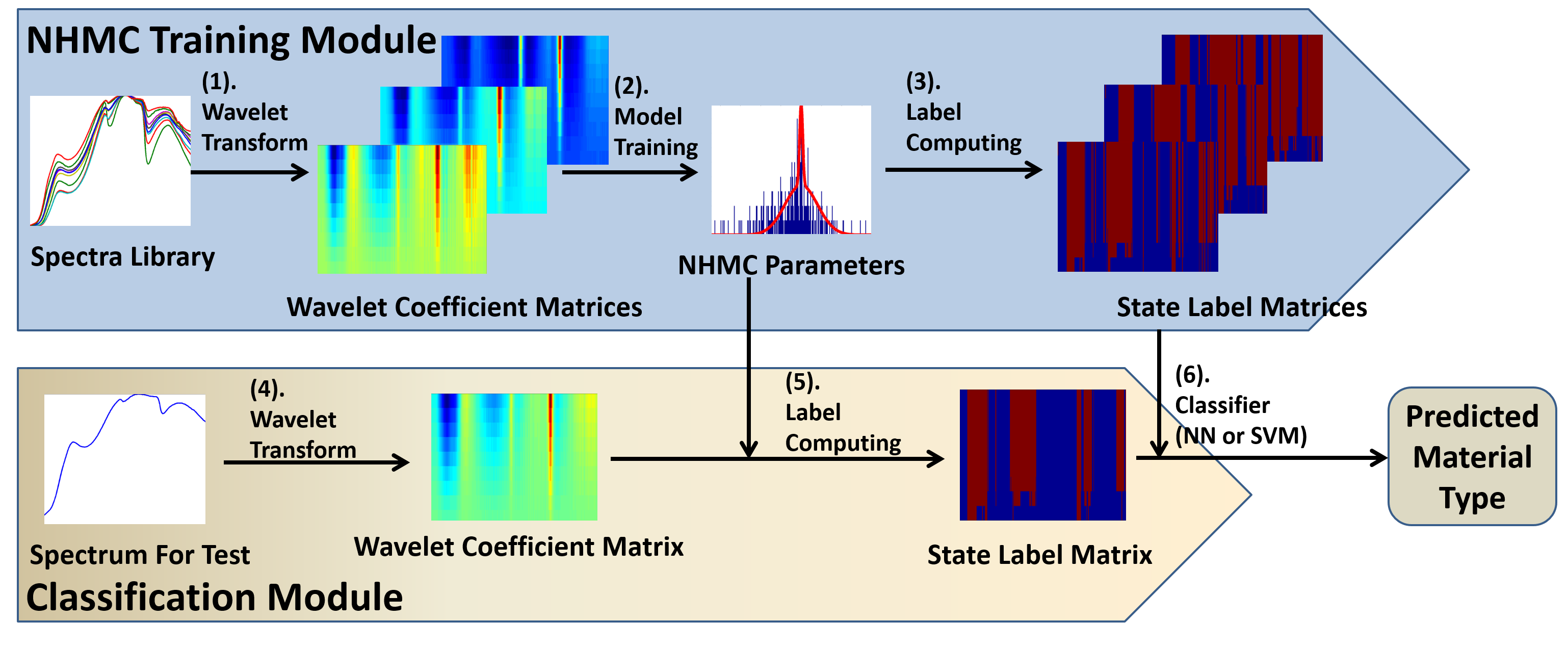}}
\caption{System overview. Top: The NHMC Training Module collects a set of training spectra, computes UWT coefficients for each, and feeds them to an NHMC training unit that outputs Markov model parameters and state labels for each of the training spectra, to be used as classification features. Bottom: The Classification Module considers a test spectrum, obtains its UWT coefficients, and extracts a state array from the NHMC obtained during training. A nearest-neighbors classifier searches for the most similar state array among the training data, and returns the class label for the corresponding spectrum.}
\label{fig:one}
\vspace{-3mm}
\end{figure*}
We provide an overview of the NHMC-based hyperspectral classification system in Fig.~\ref{fig:one}. The system consists of two modules: an NHMC model training module and a classification module. While the figure assumes a binary-state Gaussian mixture model (GMM) in the NHMC, as described in Section~\ref{sec:bgnhmc}, one can easily formulate a $k$-ary GMM state variant, $k = 2,3,\ldots$, as described in Section~\ref{sec:nhmc}. The training stage uses a training library of spectra containing samples from the classes of interest to train the NHMC model, which is then used to compute state estimates for each of the training spectra using a Viterbi Algorithm. The state arrays obtained from the NHMC model will then be used as classification features coupled with a classification scheme, e.g., nearest-neighbor (NN) or support vector machine (SVM) classification. 
The testing module considers a spectrum under test and computes the state estimates under the trained NHMC model using the parameters obtained during training. The module then applies the classification scheme being tested, returning the class label of the selected training spectrum.\par

\section{Classification Experiments and Result Analysis}
\label{sec:exps}

In this section, we present multiple experimental results that assess the performance of the proposed features in hyperspectral signature classification. We also study the effect of NHMC parameter selections on the classification performance from the corresponding extracted features.

\subsection{Study Data and Performance Evaluation}
The dataset used in this paper is a part of the RELAB spectral database with 26 mineral reflectance spectrum classes. Since the spectra in the original database have different wavelength ranges, we only use the spectral region from 0.35~\textrm{\mum} to 2.6~\textrm{\mum} (if applicable) which contains almost all of the visible and near-infrared region of the electromagnetic spectrum. We only use the spectra with spectral resolution being 5~nm to eliminate the differences in spectral resolution in different sources. A different number of samples is present in each mineral class. Thus, in order to ensure the same weight of each class in the training process, we use the Hapke mixing model~\cite{hapke2012theory} to generate additional mixtures of existing spectra in a given class until all classes have the same number of samples. We do this to prevent different mineral types from having different contributions to the model obtained and influencing the final classification accuracy. The final dataset contains 1690 reflectance spectra with each class including 65 reflectance spectra. Additionally, in order to eliminate the influences caused by illumination conditions, we perform normalization on the whole database by dividing each reflectance spectrum by its maximum value. \par
We compare different NHMC models (both GMM and MOG with different number of mixed Gaussian components and with/without assigning Haar wavelet coefficient signs to state labels). We first randomly separate the dataset into a training library (including 1352 samples with each class containing 52 reflectance spectra) and a test set (including 338 samples with each class containing 13 reflectance spectra). In order to evaluate the performance of different NHMC-based features, we train these NHMC-based features on the training library. Then we use the Viterbi algorithm to obtain the corresponding state labels for both the training library and test set and use both linear and non-linear classifiers (nearest neighbor (NN) classifier, support vector machine (SVM) classifier) on the test set to evaluate the classification accuracy of different models. \par
Unfortunately, the resulting dataset features a significant separation between the different classes, and so it is difficult to differentiate the performances of the different proposed methods, which are very high. In order to discriminate among the methods, we introduced mixing into the database as an attempt to increase the variability among reflectance spectra in each given class. Our mixing methodology is designed to resemble the image blurring process common in hyperspectral imaging. First, we randomly order the reflectance spectra in the database into a 3-D array (a so-called datacube) with two spatial dimensions and one spectral dimension. We then perform identical spatial blurring on each wavelength using a $3\times 3$-pixel Gaussian smoothing operator. Finally, we build a new library from the blurred pixels' spectra while retaining the original labels. By performing this image-based blurring, each spectrum in the resulting database exhibits a mixture of structural features from spectra in multiple classes, which provides a more challenging spectrum classification setup. We vary the Gaussian blurring kernel variance among a range of values to adjust the amount of mixing performed: the {\em dominant material percentage} (DMP) of the original pixel in the corresponding blurred pixel is obtained as the normalized weight of the central element in a Gaussian smoothing operator. In our experiment, we vary the DMP from $70\%$ to $100\%$ with a step of $5\%$. \par

\subsection{Feature Comparison}

For this study, classification performance is evaluated by using NN and SVM classification accuracies. For the NN classifier, three distance metrics are employed: $\ell_1$ distance, Euclidean ($\ell_2$) distance, and cosine similarity measure. For the SVM classifier, we use radial basis function (RBF) as the kernel and perform a grid search for the corresponding parameter values (cost and Gaussian variance) that provide best performance for each NHMC model. Both the NHMC model (if applicable) and the classifier (NN or SVM) are trained using the aforementioned training set, and the performance is measured on the aforementioned test set. 

\begin{figure*}[t]
\centering
\centerline{\includegraphics[width=8cm]{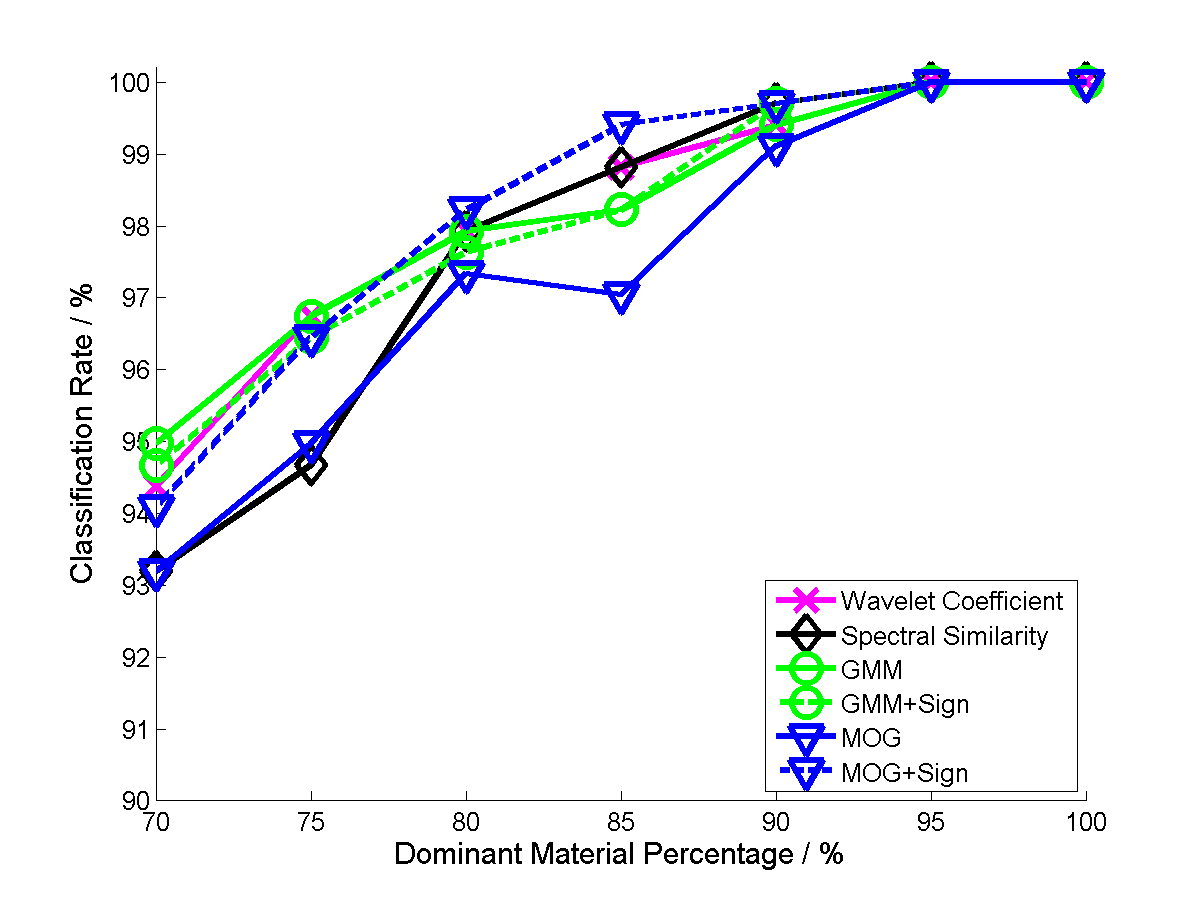}\includegraphics[width=8cm]{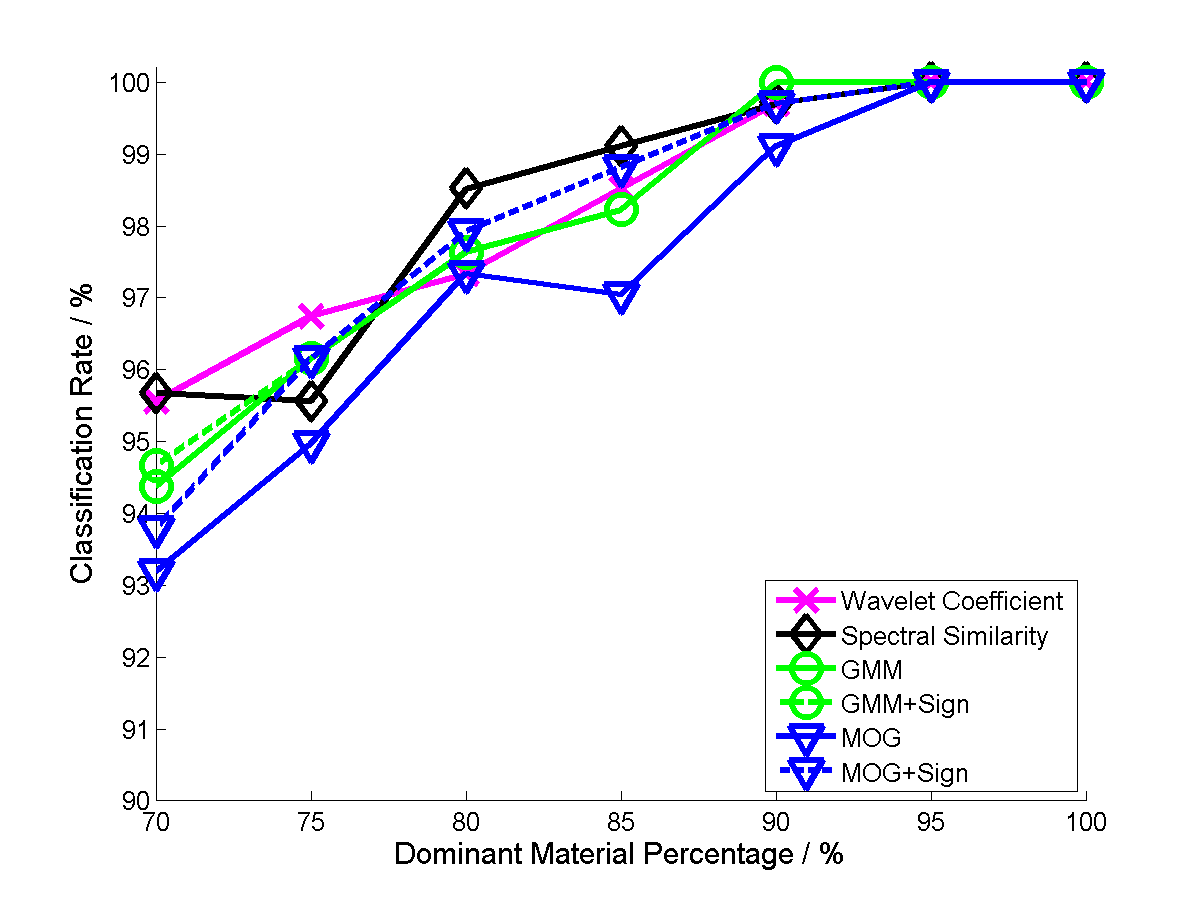}}
\centering
\centerline{\includegraphics[width=8cm]{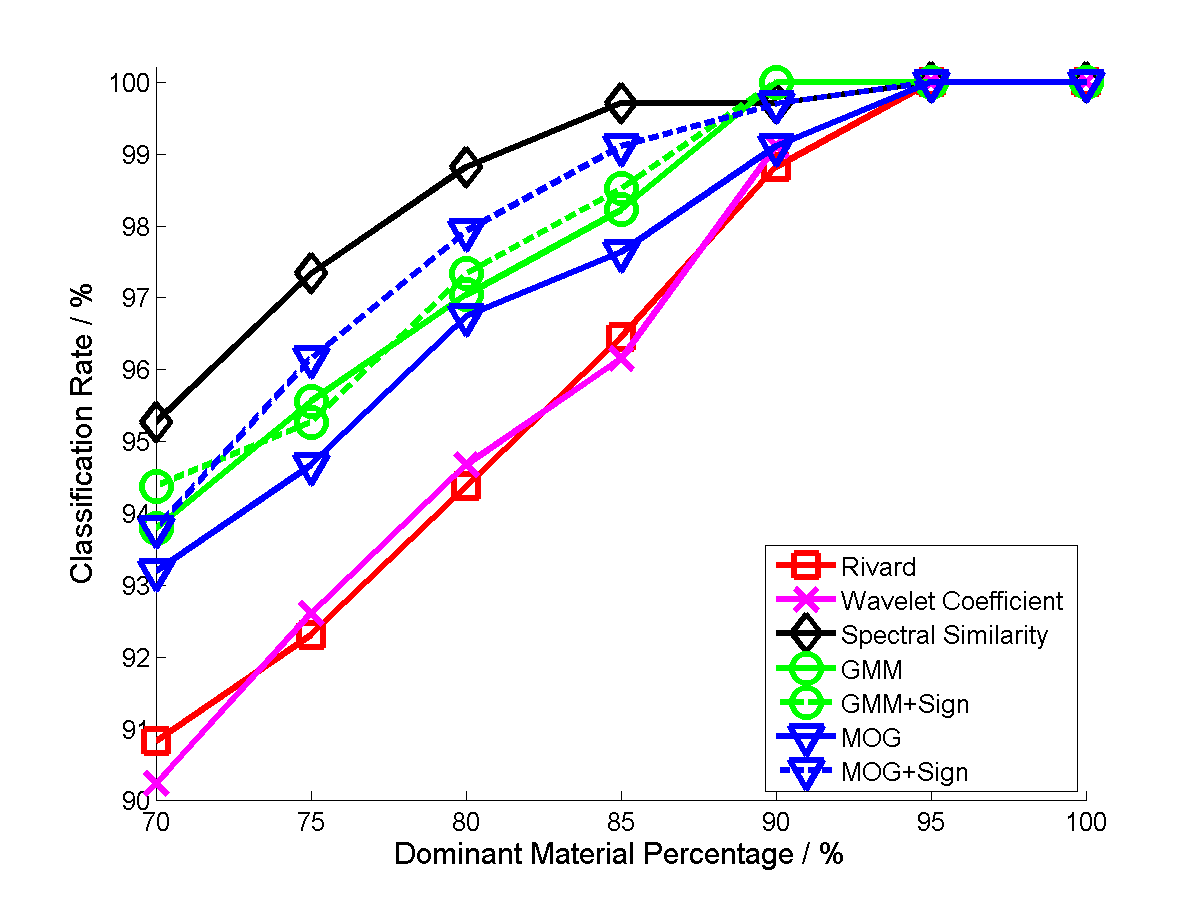}\includegraphics[width=8cm]{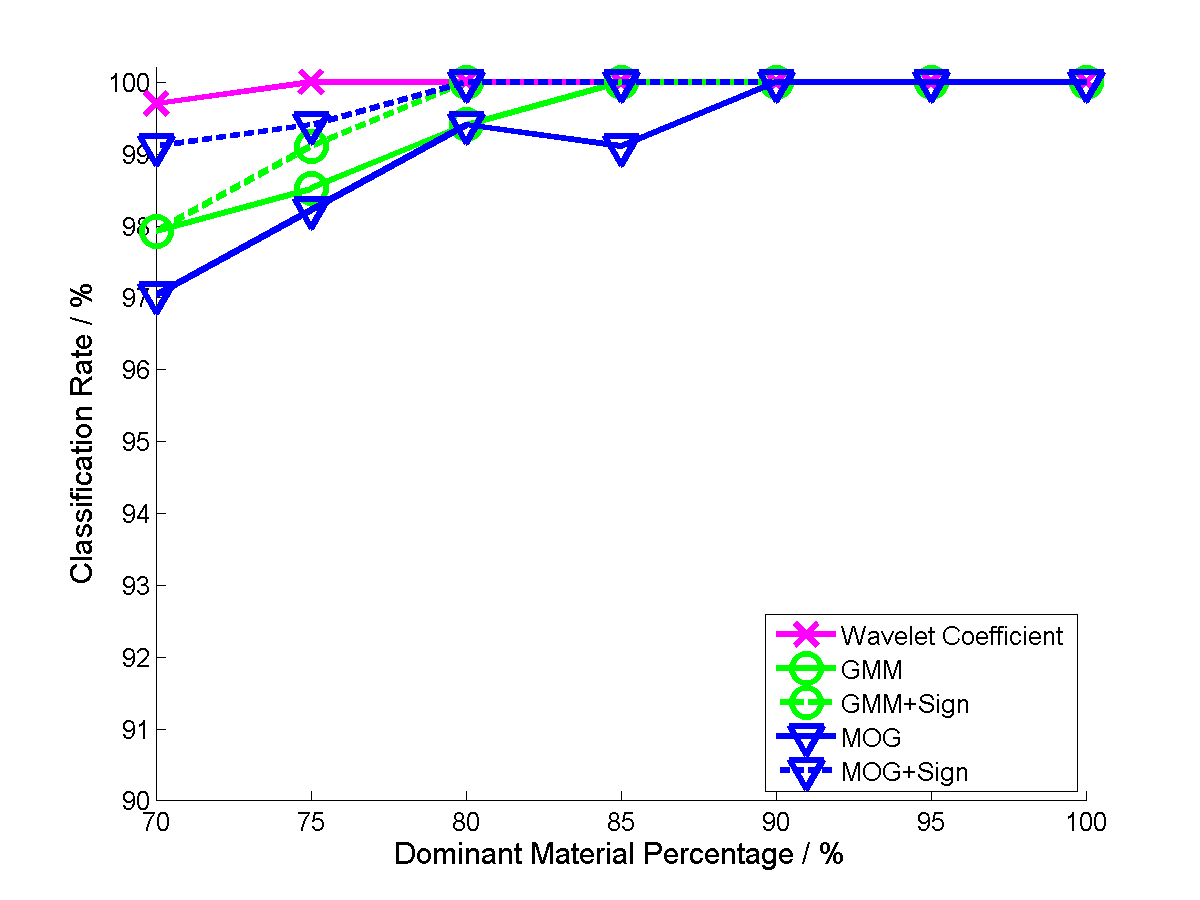}}
\caption{Classification rates of different NHMC modeling approaches and other relative classification approaches under different dominant material percentages. Top left: NN classifiers with $\ell_1$ distance; top right: NN classifier with Euclidean distance; bottom left: NN classifier with cosine similarity; bottom right: SVM classifier. For NHMC models, the highest classification rate among the models tested is listed for each DMP value. }
\label{fig:five}
\end{figure*}

Figure~\ref{fig:five} shows the classification rates for different NHMC models under different dominant material percentages using the aforementioned NN and SVM classifiers. Additionally, the figure also includes the classification accuracies of the related approaches described in Section~\ref{sec:wavspec}. In the figure, different classification features are identified as follows: ``Rivard" denotes the approach proposed in~\cite{rivard2008continuous};\footnote{Note that ``Rivard'' only appears in the bottom left figure of Fig. \ref{fig:five}  because it is defined specifically in terms of a NN classifier with cosine distance \cite{rivard2008continuous}.} ``Wavelet Coefficient" denotes the classification scheme of using wavelet coefficients as classification features; ``Spectral Similarity" denotes spectral similarity matching classification scheme (i.e., the spectra themselves are the input to each NN classifier); ``GMM" denotes an NHMC featuring Gaussian mixture models; ``MOG" denotes an NHMC featuring mixtures of Gaussians; and ``GMM+Sign" and ``MOG+Sign" denotes the previous two approaches where Haar wavelet coefficient signs being added to state labels.  Our NHMC tests involve NHMC models containing different numbers of mixed Gaussian components; Fig. \ref{fig:five} shows the highest performance among all tested values for the number of mixed Gaussian components, and Tables \ref{table:one}-\ref{table:four} list the best-performing values for each DMP. 

\begin{table*}[!t] \footnotesize 
\renewcommand{\arraystretch}{1.3}
\caption{Number of Gaussian components mixed achieving highest classification accuracy with nearest neighbor classifier in conjunction with $\ell_1$ distance.}
\label{table:one}
\centering
\begin{tabular}{|c||c||c||c||c||c||c||c|}
\hline
\backslashbox{Model}{DMP} & 70\% & 75\% & 80\% & 85\% & 90\% & 95\% & 100\%\\
\hline
GMM & 10 & 6 & 10 & 2 & 7-10 & 8-10 & 3-10\\
\hline
GMM+sign & 10 & 6 & 4,5,10 & 2 & 7-9 & 2-10 & 2-10\\
\hline
MOG & 4,7,8 & 5 & 7 & 3,5,7-9 & 5-7,9 & 6-10 & 3-10\\
\hline
MOG+sign & 4 & 10 & 3 & 10 & 9 & 3-10 & 3-10\\
\hline
\end{tabular}
\end{table*}

\begin{table*}[!t] \footnotesize 
\renewcommand{\arraystretch}{1.3}
\caption{Number of Gaussian components mixed achieving highest classification accuracy with nearest neighbor classifier in conjunction with Euclidean distance.}
\label{table:two}
\centering
\begin{tabular}{|c||c||c||c||c||c||c||c|}
\hline
\backslashbox{Model}{DMP} & 70\% & 75\% & 80\% & 85\% & 90\% & 95\% & 100\%\\
\hline
GMM & 5,10 & 5 & 10 & 2 & 9 & 6-8,10 & 3-10\\
\hline
GMM+sign & 6 & 3 & 4,9,10 & 5 & 6,9 & 2-10 & 2-10\\
\hline
MOG & 7,8 & 5 & 7 & 3,5,7-9 & 5-7,9 & 6-10 & 3-10\\
\hline
MOG+sign & 4,9 & 4,10 & 3 & 7-10 & 9 & 3-10 & 3-10\\
\hline
\end{tabular}
\end{table*}

\begin{table*}[!t] \footnotesize 
\renewcommand{\arraystretch}{1.3}
\caption{number of Gaussian components mixed achieving highest classification accuracy with nearest neighbor classifier in conjunction with cosine similarity.}
\label{table:three}
\centering
\begin{tabular}{|c||c||c||c||c||c||c||c|}
\hline
\backslashbox{Model}{DMP} & 70\% & 75\% & 80\% & 85\% & 90\% & 95\% & 100\%\\
\hline
GMM & 6 & 6 & 5,9,10 & 4 & 5,9 & 6,7,9,10 & 3-10\\
\hline
GMM+sign & 9 & 6,9 & 3,5 & 5 & 5,7 & 2-10 & 2-10\\
\hline
MOG & 4 & 4 & 7 & 8 & 5-7,9 & 6-10 & 3-10\\
\hline
MOG+sign & 3,5 & 5 & 3 & 6-8 & 9 & 3-10 & 3-10\\
\hline
\end{tabular}
\end{table*}

\begin{table*}[!t] \footnotesize 
\renewcommand{\arraystretch}{1.3}
\caption{Number of Gaussian components mixed achieving highest classification accuracy with support vector machine classifier in conjunction with radial basis function.}
\label{table:four}
\centering
\begin{tabular}{|c||c||c||c||c||c||c||c|}
\hline
\backslashbox{Model}{DMP} & 70\% & 75\% & 80\% & 85\% & 90\% & 95\% & 100\%\\
\hline
GMM & 8 & 10 & 4,7-10 & 3 & 4-10 & 3-10 & 2-10\\
\hline
GMM+sign & 2 & 10 & 2 & 3,5,8,9 & 4-10 & 2-10 & 2-10\\
\hline
MOG & 6 & 7 & 6,7,9,10 & 9 & 4 & 5-10 & 3-10\\
\hline
MOG+sign & 3 & 3,5-7 & 4,5 & 3-10 & 3-10 & 3-10 & 3-10\\
\hline
\end{tabular}
\end{table*}

We highlight some features of the obtained results:
\begin{itemize}
\item In most cases, the use of signs in the NHMC features improves performance with respect to their original counterparts.
\item In the NN classifiers, GMM performs better than MOG for lower DMPs, which are more challenging settings, while MOG with additional signs outperforms GMM for DMPs closer to 100\%. Nonetheless, in most cases MOG without wavelet coefficient signs provides the worst performance.
\item While the performance of NHMC methods with SVM classifiers is higher than that obtained with NN classifiers, they are outperformed by the wavelet coefficient approach. We conjecture that this is due to the discrete nature of NHMC labels, which are not as easily leveraged in the SVM's search for a separation boundary from support vectors.
\end{itemize}
We also attempted to implement the approach proposed in~\cite{prasad2012information}. However, because of the lack of sufficient data for individual classes, we obtained several ill-conditioned covariance matrices when constructing multivariate GMMs. Thus, we do not include the comparison with this approach in this paper.\par

\subsection{NHMC Parameters}

\begin{figure*}[t]
\centering
\centerline{\includegraphics[width=8cm]{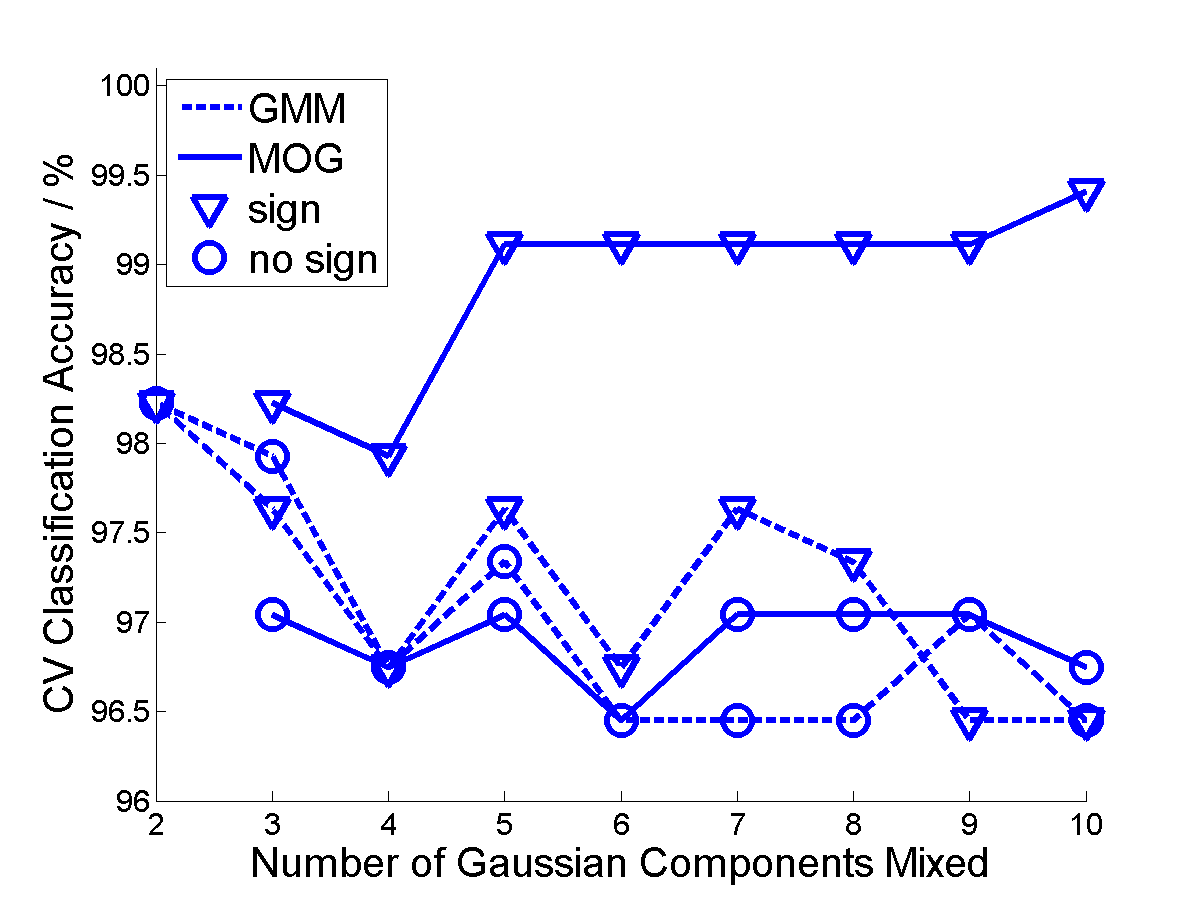}\includegraphics[width=8cm]{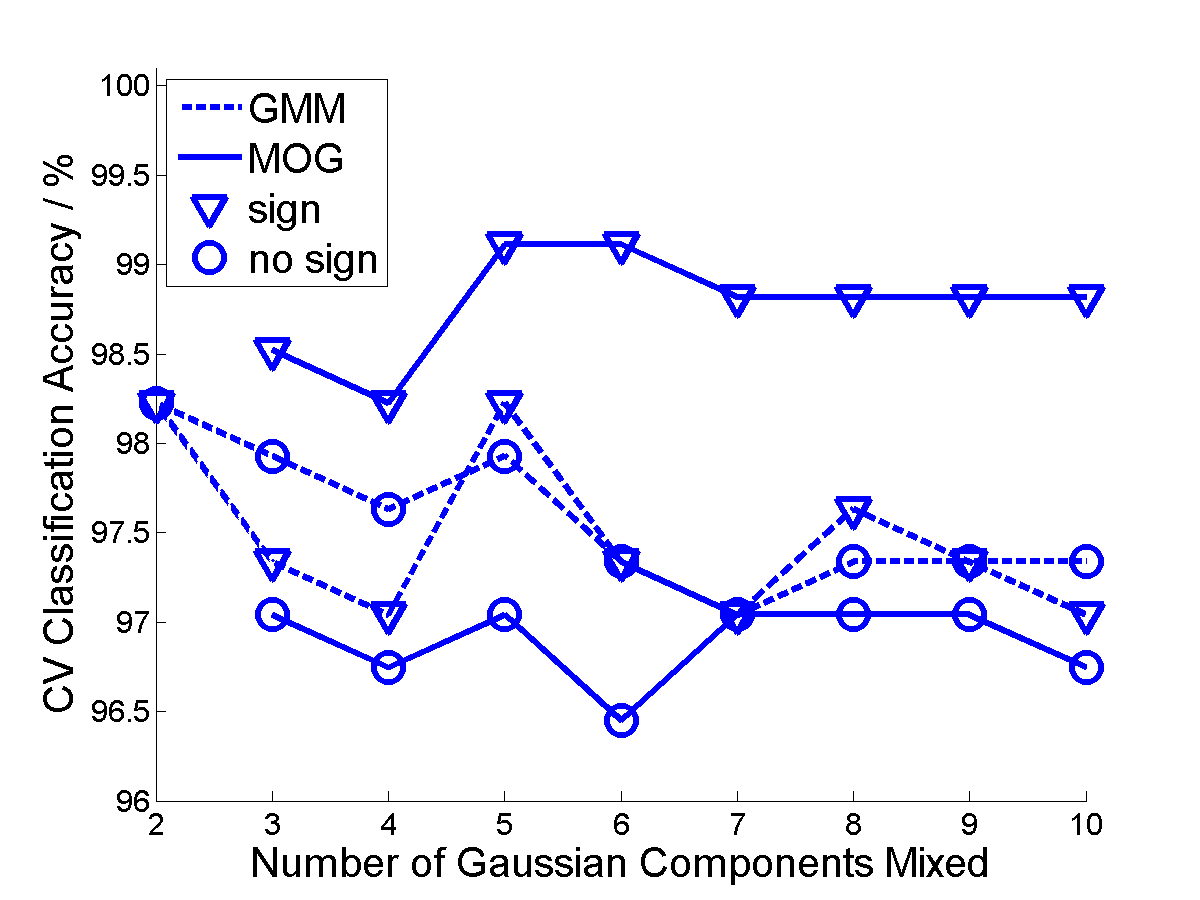}}
\centering
\centerline{\includegraphics[width=8cm]{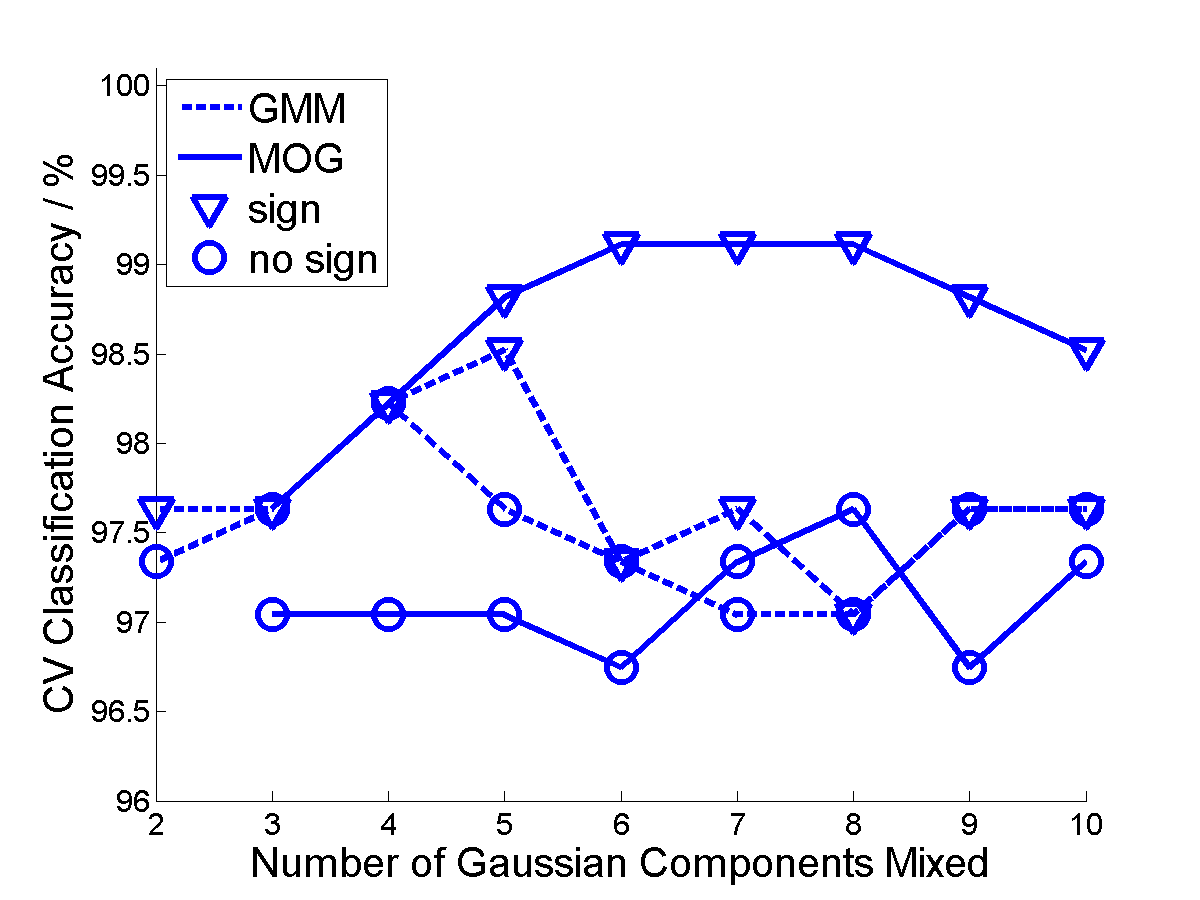}\includegraphics[width=8cm]{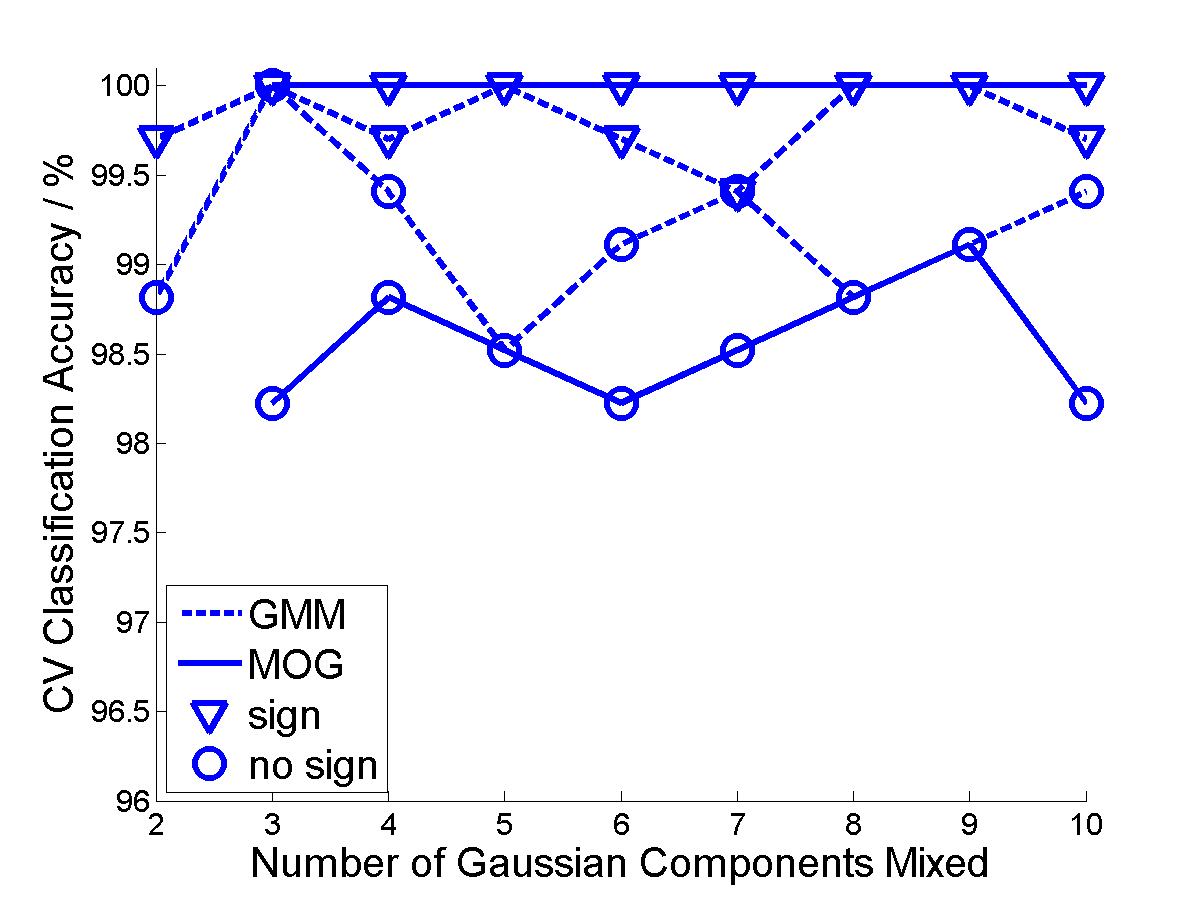}}
\caption{Classification rates of using different number of mixed Gaussian states with DMP of 85\%. Top left: NN classifiers with $\ell_1$ distance; top right: NN classifier with Euclidean distance; bottom left: NN classifier with cosine similarity; bottom right: SVM classifier.}
\label{fig:six}
\end{figure*}

Next, we evaluate the effect of the number of states included in the NHMC model on the performance of the tested classifiers. We set the DMP to 85\% for concreteness, and evaluate the classification performance of all proposed NHMC features with NN and SVM classifiers as a function of the number of states, which varies between 2 and 10 for GMM and between 3 and 10 for MOG. Fig.~\ref{fig:six} shows the variation tendency of classification accuracy with increasing number of mixed Gaussian components using different classifiers and similarity metrics.\par
From these four figures, we see that MOG with additional wavelet coefficient signs provides relatively consistent performance compared with other NHMC-based models. Additionally, in terms of classification accuracy, the two model configurations using MOG provide two performance extremes: by adding wavelet coefficient signs we obtain the highest classification performance, while MOG without signs provides the lowest one. As mentioned earlier, MOG combines the simplicity of a binary-state GMM and the spectral fluctuation characterization capability of a multistate GMM. In that case, if we do not consider the signs of the wavelet coefficient, spectra that have approximately matching locations for their fluctuations while exhibiting differing magnitudes and orientations will be matched to similar MOG label vectors. The reason is that a binary-state GMM form could assign the same state labels to several fluctuations of different levels and orientations. However, if Haar wavelet coefficient signs are added, the state labels better reflect the spectral fluctuation orientation information.\par

\section{Conclusion}
\label{sec:conc}

In this paper, we proposed the design of a feature extraction scheme for hyperspectral signatures that preserves the semantic information used by practitioners in signature discrimination  (i.e., location of distinguishing fluctuations and discontinuities). Our approach is automated thanks to the use of statistical models for wavelet transform coefficients, which succinctly capture the location and magnitude of fluctuations in the spectra observed. Furthermore, the statistical model also enables a segmentation of the spectra into informative and non-informative portions. The success of statistical modeling is mostly dependent on the availability of a large-scale database for training containing representative examples of the spectra that are observed by the sensing system.

We also tested the quality of the preservation of semantic information in our proposed features by using a simple example hyperspectral classification system based on nearest neighbor search. We also compared our feature extraction method with three existing feature extraction approaches for classification; the first approach is spectral matching, which performs classification directly on the hyperspectral signature; the second approach performs classification directly on wavelet coefficients, and the third approach computes features as the sum of wavelet coefficients of certain scales. We showed that the performance of our proposed features meets or exceeds that of baselines relying on spectral matching and wavelet coefficient representations, in particular for high DMP. 

While the performance of each method we tested decreases as the DMP is reduced, the reduction is stronger for the MOG and GMM methods in comparison with some of their counterparts (in particular, to the case where NN with cosine similarity is applied directly on the spectra). We believe that this effect is due to the additional difficulty of modeling signal classes of increased variability (as the DMP decreases) using the extracted binary features. Nonetheless, we note that even with this handicap the performance of the best combinations of NHMC features and NN classifiers exceeds the performance of the comparison baseline methods when the DMP is sufficiently large. Furthermore, we believe that the size of the datasets we use here, while much larger than that of our previous results~\cite{MParente:a,MFDuarte:non-homogeneous,SFeng:tailoring}, may still be insufficient to fully exploit the power of the statistical models leveraged here. Thus, further work will focus on expanding the size of database and investigating additional modifications to the feature extraction scheme and the underlying statistical models. As an example, NHMC models based on nonzero-mean GMM are an attractive alternative to be pursued in the future, as in certain cases the histogram of wavelet coefficients cannot be accurately modeled by zero-mean Gaussian mixture models.

\section*{Acknowledgements}

We thank Mr. Ping Fung for providing an efficient implementation of our NHMC training code that largely decreased the time of running experiments. 

\begin{appendices}
\section{Proof of Lemma~\ref{lemma:one}}
\label{app:lemmaone}
By denoting
$p_{a \rightarrow b}^{s, S \rightarrow Z}=p(Z_{s}=b|S_{s}=a)$
and using law of total probability, we can get
$p(Z_{s}=b)=\sum_{a} p_{a \rightarrow b}^{s, S \rightarrow Z}p(S_{s}=a)$.
From the $Z(S)$ map in eq.\ (\ref{eq:map}), we can infer that
$p_{0 \rightarrow 0}^{s, S \rightarrow Z}=1$ and
$p_{i \rightarrow 1}^{s, S \rightarrow Z}=1$.
Therefore, it is easy to derive the conclusion in Lemma~\ref{lemma:one}.

\section{Proof of Lemma~\ref{lem:mog}}
\label{app:mog}
\begin{figure}[t]
\centering
\centerline{\includegraphics[width=5cm]{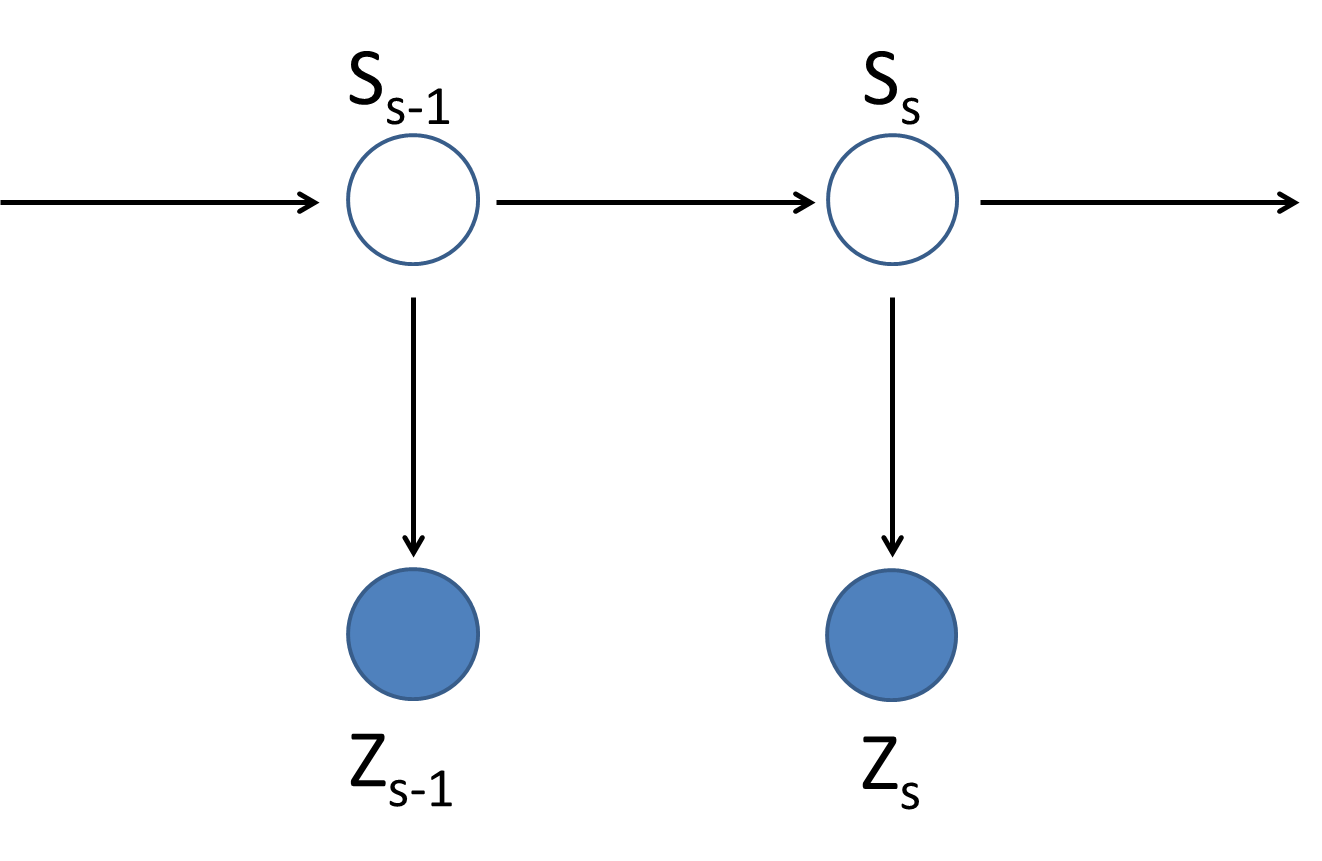}}
\caption{Directed graphic model for MOG and GMM state labels.}
\label{fig:seven}
\end{figure}

The relationship between the original state labels $S_{s-1}$, $S_{s}$ and the combined state labels $Z_{s-1}$, $Z_{s}$ can be characterized by a directed graphical model shown in Fig. \ref{fig:seven}. 
By considering all possible transitions from $Z_{s-1}$ to $Z_{s}$ through the state transitions $S_{s-1}$ to $S_s$ and the map above, and denoting 
\begin{align*}
p_{a \rightarrow b}^{s, Z \rightarrow S}&=p(S_{s}=b|Z_{s}=a),
\end{align*}
we appeal to the law of total probability to write
\begin{equation}
q_{b\rightarrow a,s}=\sum_{x=0}^{k-1} \sum_{y=0}^{k-1} p_{x\rightarrow a}^{s, S \rightarrow Z} p_{y \rightarrow x,s} p_{b \rightarrow y}^{s-1, Z \rightarrow S}.
\label{eq:chain}
\end{equation}
From the $Z(S)$ map in equation (\ref{eq:map}), we can also infer that
$p_{0 \rightarrow 1}^{s, S \rightarrow Z}=0$,
$p_{i \rightarrow 0}^{s, S \rightarrow Z}=0,$
$p_{0 \rightarrow 0}^{s-1, Z \rightarrow S}=1$,
$p_{0 \rightarrow i}^{s-1, Z \rightarrow S}=0$,
$p_{1 \rightarrow 0}^{s-1, Z \rightarrow S}=0$, and

\begin{align*}
p_{1 \rightarrow i}^{s-1, Z \rightarrow S}&=\frac{p(S_{s-1}=i)}{\sum_{j=1}^{k-1} p(S_{s-1}=j)},
\end{align*}
where $i=1,...,k-1$. After combining the equalities above with $(\ref{eq:chain})$ for $a,b \in \{0,1\}$, we can get the four elements in new matrices expressed in $(\ref{eq:lem1}-\ref{eq:lem4})$, proving the lemma.

\end{appendices}


\end{document}